# ROBUST AND SCALABLE LEARNING OF COMPLEX DATASET TOPOLOGIES VIA ELPIGRAPH

Luca Albergante[1,2,3*], Evgeny M. Mirkes[4,5], Huidong Chen[6,7,8], Alexis Martin[1,2,3,10], Louis Faure[1,2,3,9], Emmanuel Barillot[1,2,3], Luca Pinello[6,11], Alexander N. Gorban[4,5], Andrei Zinovyev[1,2,3*]

[1] Institut Curie, PSL Research University, F-75005 Paris, France

[2] INSERM, U900, F-75005 Paris, France

[3] MINES ParisTech, PSL Research University, CBIO-Centre for Computational Biology, F-75006 Paris, France

[4] Department of Mathematics, University of Leicester, University Road, Leicester LE1 7RH, UK

[5] Lobachevsky University, Nizhni Novgorod, Russia

[6] Molecular Pathology Unit & Cancer Center, Massachusetts General Hospital and Harvard Medical School, Boston, MA 02114, USA

[7] Department of Biostatistics and Computational Biology, Dana-Farber Cancer Institute and Harvard T.H. Chan School of Public Health, Boston, MA 02215, USA

[8] Department of Computer Science and Technology, Tongji University, Shanghai 201804, China

[9] Institute of Technology and Innovation (ITI), PSL Research University, PSL Research University, F-75005 Paris, France

[10] ECE Paris, F-75015 Paris, France

[11] Broad Institute of MIT and Harvard, Cambridge, MA 02142, USA

*The correspondence should be sent to Andrei.Zinovyev@curie.fr or Luca.Albergante@curie.fr .

*Abstract (179)*

Large datasets represented by multidimensional data point clouds often possess non-trivial distributions with branching trajectories and excluded regions, with the recent single-cell transcriptomic studies of developing embryo being notable examples. Reducing the complexity and producing compact and interpretable representations of such data remains a challenging task. Most of the existing computational methods are based on exploring the local data point neighbourhood relations, a step that can perform poorly in the case of multidimensional and noisy data. Here we present ElPiGraph, a scalable and robust method for approximation of datasets with complex structures which does not require computing the complete data distance matrix or the data point neighbourhood graph. This method is able to withstand high levels of noise and is capable of approximating complex topologies via principal graph ensembles that can be combined into a consensus principal graph. ElPiGraph deals efficiently with large and complex datasets in various fields from biology, where it can be used to infer gene dynamics from single-cell RNA-Seq, to astronomy, where it can be used to explore complex structures in the distribution of galaxies.

**Introduction (1444 words)**

Modern "big data" datasets are frequently characterized by complex structures, which are difficult to appreciate by simple data visualization and data approximation methods. For example, recently obtained snapshot distributions of single cells of a developing embryo or an adult organism, when looked in the space of their transcriptomic profiles, display complex *data point clouds* characterized by branching or converging (i.e., forming loops) developmental trajectories, regions of varying local dimensionality, and high level of biological and technical noise[1–3]. To better characterize and quantify the structure of such dataset, it is important to develop computational methods aimed at providing a low complexity data representation, while preserving some essential features of the multidimensional data distribution.

When considering data distributions in high-dimensional spaces, two opposed scenario need to be kept in mind[4,5]. In some cases, clouds of data points can be localized in the vicinity of a relatively low-dimensional object (such as a principal manifold), and hence possess a *low intrinsic dimensionality*. Under these circumstances, numerous dimensionality reduction approaches currently used can be efficient in recovering the low-dimensional object, either explicitly or implicitly, and in projecting the data points onto it. This is the case of high extrinsic but low intrinsic dimensionality, where an informative low dimensional data projection exists. However, some clouds of data points are characterized by a truly high-dimensional structure. In this case, mathematical phenomena such as concentration of measures start to play an important role, and dimensionality reduction methods can become inadequate[6]. Nonetheless, such *curse of dimensionality* can also be a *blessing*, and approaches based on self-averaging or on the application of stochastic separability theory can be very successful[7,8].

Manifold learning methods aim at modelling the multidimensional data as a noisy sample from an underlying generating manifold, usually of relatively small dimension. The noise present in the sample has a double nature. *Sampling noise* scatters the data points around the generating manifold in a relatively close vicinity, while *background noise* introduce points generated independently from the manifold (which can be considered bona fide outliers). A classical linear manifold learning method is Principal Component Analysis (PCA) introduced more than 100 years ago[9]. From the 1990s, multiple generalizations of PCA to non-linear manifolds have been suggested, including Self-Organizing Maps (SOMs)[10], elastic maps[11,12], ISOMAP[13], Local Linear Embedding (LLE)[14], t-distributed stochastic neighbor embedding[15], regularized principal curves and manifolds[16], UMAP[17] and many others[18]. In certain contexts, the rigorous definition of manifold may be too restrictive and it might be advantageous to construct data approximators in the form of more general mathematical varieties by gluing manifolds on their boundaries, thus introducing singularities which can correspond, for example, to branching points. Simplicial complexes (i.e., sets composed of points, line segments, triangles, and n-dimensional simplexes) can provide a basis for constructing rather generic data approximators[19]. Moreover, they can reflect the non-trivial topology to be discovered in the data and account for varying local dimensionality. However, we still lack tractable methods for robust extraction of such general objects from the data. Currently, the most used non-manifold type data approximators are principal graphs[20], with principal trees as the simplest and most tractable graph type. Principal graphs are data approximations constructed by graph embedding "passing through the middle of data" and possessing specific regular properties, which restrict the graph complexity[20,21] (see Figure 1A and the formal definition below).

High demand of appropriate methods for principal graph reconstruction has emerged recently in connection with novel sequencing technologies in molecular biology which often generate high-dimensional datasets characterized by higher level of geometrical complexity. For example, clouds of data points representing heterogeneity of transcriptomic profiles of thousands of single cells are frequently characterized by curvilinear and branching structures. These structures reflect continuous changes in the regulatory programs of the cells and their bifurcations during complex cell fate decisions. The existence and biological relevance of such branching trajectories was clearly demonstrated while studying development[22], cellular differentiation[23–25] and cancer biology[26]. The power of such analyses stimulated the emergence of a number of tools for reconstructing so called "cellular trajectories" and "branching pseudotime" in the field of bioinformatics[27–29]. Some of these tools exploit the notion of principal curves or graphs explicitly[30,31], while others sometimes use a different terminology closely related to principal graphs or principal trees in their purpose[32,33]. Biology is, however, only one of the possible domains of applicability of principal graphs. They can serve as useful data approximations in other fields of science such as political sciences or image processing[12,34].

Most of the methods currently used to learn complex non-linear data approximators are based on an auxiliary object called *k*-nearest neighbour (*k*NN) graph, constructed by connecting each data point to its *k* closest (in a chosen metrics) neighbouring points (Figure 1B). To avoid unnecessary complexity, the *k*NN graph, or a similar object, can be constructed using pre-clustered data distribution or a sample of the data[35]. A popular mathematical tool for extracting the approximator graph structure from the *k*NN graph is the Minimal Spanning Tree (MST) or computationally feasible heuristics for its estimation. Despite its popularity, using MST or similar approaches introduces certain limitations in the resulting data approximators. For example, if background noise is present or the underlying manifold spans many dimensions in the data space, the structure of the *k*NN graph, together with the reconstructed MST, can easily become very complex, non-robust and even misleading (see Figure 1B). In practice, the majority of the methods used for extracting branching data structures require drastic dimension reduction (to 3D or 2D), as the properties of *k*NN graph are more stable and tractable in such low dimensional spaces[36,37]. However, it is easy to imagine toy examples describing, for example, a tree-like manifold embedded in a multi-dimensional space, containing intersecting branches in any linear 2D projection. Projecting in 3D should make exact intersection of manifold branches less probable: nevertheless, branches might easily appear much closer, after projection into lower dimension, than they are in higher dimensions and effectively intersect (especially when a manifold is sampled with noise). Moreover, most of the methods described in the literature rely on heuristics for estimating the optimal graph structure (such as the MST, for tree-like topologies) and do not explore sufficiently large volume of the structural space to determine which approximating graph topology best describes the data.

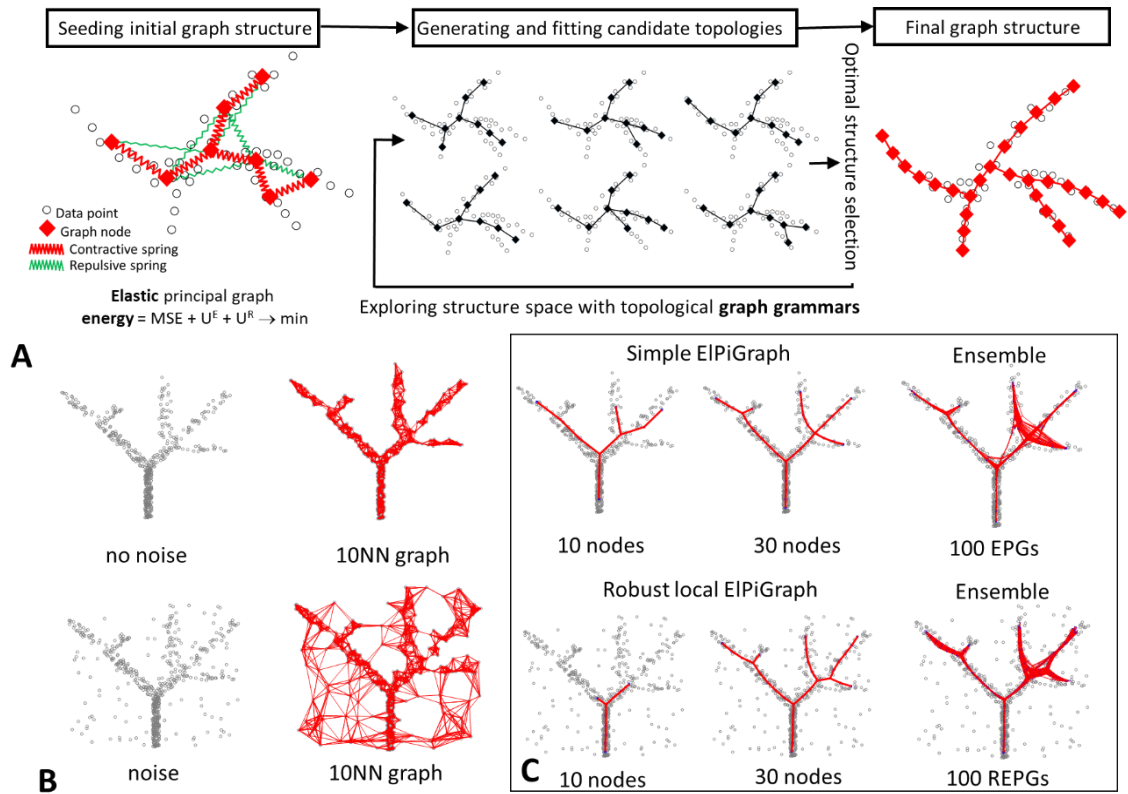

**Figure 1. Basic principles and examples of ElPiGraph usage. (A)** Schematic workflow of the ElPiGraph method. Left, construction of the elastic graph starts by defining the initial graph topology and embedding it into the data space. The graph structure is fit to the data, using minimization of the mean square error regularized by the elastic energy. The elastic energy includes a term reflecting the overall stretching of the graph (symbolically shown as contractive red springs) and a term reflecting the overall bending of the graph branches and the harmonicity of branching points (shown as repulsive green springs). Middle, ElPiGraph explores a large region of the structural space by exhaustively applying a set of graph rewriting rules (graph grammars) and selecting, at each step, the structure leading to the minimum overall energy of the graph embedding. **(B)** The structure of the $k$NN graph, currently used in many manifold learning methods, can be misleading in the case of high dimensional data, when data are sampled with noise from a generating manifold, or when background noise is present, as illustrated here. **(C)** Left and middle, illustration of the robust local workflow of ElPiGraph, which makes it possible to deal with the presence of noise. In the global version, the graph structure is fit to all data points at the same time, while in the local version, the structure is fit to the points in the local graph neighbourhood, which expands as the graph grows. Right, an illustration of principal graph ensemble approach: 100 elastic principle graphs are superimposed, each constructed on a fraction of data points randomly sampled at each execution.

In this paper we present a method for constructing elastic principal graphs together with its algorithmic implementation which we named ElPiGraph. The ElPiGraph method does not employ the notion of $k$NN graph, MST, or data pre-clustering. It does not require data preprocessing via drastic dimensionality reduction and does not require the construction of the complete distance matrix between data points. The core algorithm used by ElPiGraph to fit a fixed graph structure to the data is almost as fast as standard $k$-means clustering. This speed allows us to use it for exploring many graph topologies and finding the optimal one by a gradient descent-like approach (via the application of *graph grammars*) in the space of graph structures. Moreover, ElPiGraph is able to explore complex data structures via *principal graph ensembles* and to build *consensus principal graph* from such ensembles. Furthermore, ElPiGraph can be made extremely robust with respect to the presence of background noise, without affecting its computational performance. These properties make ElPiGraph a highly competitive method among existing analogues pursuing similar objectives[29].

ElPiGraph combines several concepts previously discussed, including elastic energy functional and graph grammars[11,38,39]. However, it improves significantly over existing methodology used to construct elastic principal graphs, rendering it capable to effectively approach real-word applications (such as the analysis of single cell data). ElPiGraph introduces several innovations and improvements: scalability, robustness to noise (both sampling and background), ability to construct non tree-like data approximators, and direct control over the complexity of the principal graph. To achieve this result, we designed de-novo the method for constructing elastic principal graphs by introducing an elastic matrix and a Laplacian-based penalty (see Methods section). This allowed us to scale the performance of the algorithm to a very large number of points in relatively high (~100) dimensions, which arguably makes ElPiGraph the fastest currently available method for finding and fitting complex graph topology to data. Furthermore, a set of new application-specific graph rewriting rules was introduced, allowing more focused exploration of graph space and the ability to perform optimization of the graph structure. Notably, ElPiGraph is currently implemented in five programming languages (R, Matlab, Python, Scala, Java), which makes it easily applicable across scientific domains.

In the present work, we demonstrate the ability of ElPiGraph to deal with large and complex datasets, using several synthetic and challenging real-life datasets, such as snapshot single cell transcriptomics of a developing embryo and the distribution of the galaxies in the visible Universe.

**Results (2711 words)**

*ElPiGraph is a general method for robust approximation of datasets with complex topologies*

ElPiGraph (ELastic PrIncipal Graph) is a flexible and general method for constructing and assessing confidence of data approximators having non-trivial features such as branching points and loops. Elastic principal graph approximator represents an embedment of a graph into a multidimensional space that minimizes the mean-squared distance between its nodes and the data points and, at the same time, minimizes a penalty term reflecting the complexity of the graph and its embedment map[12,20]. The core of ElPiGraph is an algorithm taking as input a finite set of data points in $R^N$ and a graph structure (i.e., a set of nodes and edges). The algorithm is able to find a mapping of graph nodes into $R^N$ that optimizes a function specifying a balance between the approximation accuracy and the mapping complexity.

The main challenge of approximating a dataset characterized by a complex topology with a principal graph is finding the optimal graph structure matching the underlying data structure. In order to do that, ElPiGraph starts with an initial guess of the graph structure and embedment, and apply a set of pre-defined rewriting rules, called *graph grammar*, in order to explore the space of all possible graph structures via a gradient descent-like algorithm and select the locally optimal one (Figure 1). One of the graph grammars allows exploring the space of possible tree-like graph topologies, but changing the grammar set allows simpler or more complex graph to be derived.

The result of ElPiGraph can be affected by outlier data points located far from the representative part of the data point cloud. In order to deal with this, ElPiGraph exploits a *trimmed approximation error* which essentially makes data points located farther than the trimming radius ($R_0$) from any graph node invisible to the optimization procedure. However, when growing, the principal graphs can gradually capture new data points, being able to learn a complex data structure starting from a simple small fragment of it. Such "from local to global" approach allows achieving great robustness and flexibility of the approximation; for example, it allows solving the problem of self-intersecting manifold clustering.

Also, the final structure of the principal graphs can be sensitive to the non-essential particularities of local configurations of data points, especially in those areas where the local intrinsic data dimensionality becomes greater than one. In order to limit the effect of a finite data sample and estimate the confidence of inferred graph features (e.g, branching points, loops), ElPiGraph applies a principal graph ensemble approach by constructing multiple principal graphs after subsampling the dataset. Posterior analysis of the principal graph ensemble allows assigning confidence scores for the topological features of the graph (i.e., branching points) and confidence intervals on their locations. The properties of principal graph ensemble can be further recapitulated into a

*consensus principal graph* possessing much more robust properties than any individual graph from the ensemble. In addition, ElPiGraph allows explicit control of the graph complexity via penalizing the order of branching points. Detailed description of the ElPiGraph method can be found in the Methods section as well as in the Supplementary Information text.

*Approximating complex data topologies in synthetic examples*

As a first step in in our analysis, we applied ElPiGraph to a synthetic 2D dataset describing a circle connected to a branching structure. ElPiGraph was able to easily recover the target structure when the approriate grammar rules are specified (Figure 2B). We underline that the use of grammars gives ElPiGraph great flexibility in approximating datasets with non-trivial topologies. In the simplest case it requires to know *a priori* the type of the topology (e.g., curve, circular, tree-like) of the dataset to approximate. However, as discussed later, ElPiGraph is also able to construct an appropriate data approximator when no information is available on the underlying topology of the data.

*Robustness of ElPiGraph to data noise*

We then explored the robustness of ElPiGraph to down- and oversampling. For a simple benchmark example of a branching data distribution in 2D (Figure 2A), we selected a fraction of data points (15%) or generated 20x more points around each of the existing ones in the original dataset. Our algorithm is able to properly recover the structure of the data, regardless of the condition being considered, with only minimal differences in the position of the branching points due to the loss of information associated with downsampling. Then, we tested the robustness of ElPiGraph to the presence of uniform background noise covering the branching data pattern (Figure 2C). As the percentage of noisy points increases, certain features of the data are not captured by the approximator, as expected. However, even in the presence of a striking amount of noise, ElPiGraph is capable of correctly recovering (at least partly) the underlying data distribution. Note that, in the examples of Figure 2C, the tree is being constructed starting from the densest part of the data space.

*Clustering intersecting manifolds via ElPiGraph*

Disentangling intersecting curvilinear manifolds is a hard to solve problem (sometimes called the *Travel Maze problem*) that has been described in different fields, particularly in computer vision[40]. The intrinsic rigidity of elastic graph in combination with a trimmed approximation error enables ElPiGraph to solve the Travel Maze problem quite naturally. This is possible because the local version of the elastic graph algorithm is characterized by certain persistency in choosing the direction of the graph growth. Indeed, when presented with a dataset corresponding to a group of three threads intersecting on the same plane, ElPiGraph is able to recognize them and cluster the data points accordingly, hence associating different paths to the different threads (Figure 2D).

*Exploiting the full dimensionality of the data*

As demonstrated by the widespread use of tools like PCA, tSNE, LLE, or diffusion maps, dimensionality reduction can be a powerful tool to better visualize the structure of data. However, some data features can be lost as the dimension of the dataset is being reduced. This can be particularly problematic if dimensionality reduction is used as an intermediate step in a long analytical pipeline. To illustrate this phenomenon, we generated a 10-dimensional dataset describing an *a priori* known branching processes. When the points are projected into a 2D plane induced by the first two principal components, one of the branches collapses and becomes indistinguishable (Figure 2E). This effect is due to the branch under consideration being essentially orthogonal to the plane formed by the first two principal components. As expected, ElPiGraph is unable to capture the collapsed branch, when it is used on the 2D PCA projection of the data. However, the branch is correctly recovered when ElPiGraph is run using all 10 dimensions.

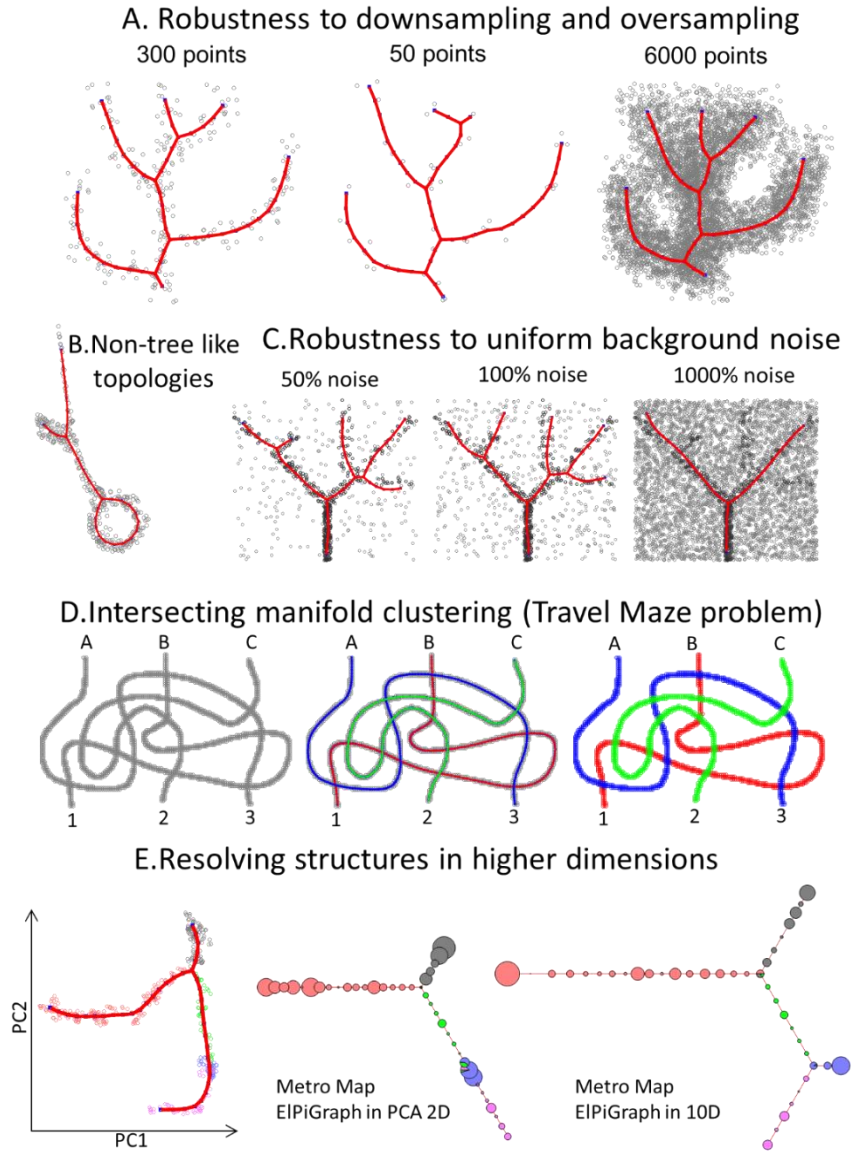

**Figure 2. Toy examples showing features of ElPiGraph.** (**A**) ElPiGraph is robust with respect to downsampling and oversampling of a dataset. Here, a reference branching dataset[31] is downsampled to 50 points (middle) or oversampled by sampling 20 points randomly placed around each point of the original dataset. (**B**) ElPiGraph is able to capture non-tree like topologies. Here, the standard set of principal tree graph grammars was applied to a graph initialized by four nodes connected to a ring. (**C**) ElPiGraph is robust to large amounts of uniformly distributed background noise. Here, the initial dataset from Figure 1 is shown as black points, and the uniformly distributed noisy points are shown as grey points. ElPiGraph is blindly applied to the union of black and grey points. (**D**) ElPiGraph is able to solve the problem of learning intersecting manifolds. On the left, a toy dataset of three uniformly sampled curves intersecting in many points is shown. ElPiGraph starts by learning a principal curve using the local version several times, each time on a complete dataset. However, for each iteration, ElPiGraph is initialized by a pair of neighbouring points not belonging to points already captured by a principal curve. The fitted curves are shown in the middle of the point distribution by using different colors, and the clustering of the dataset based on curve approximation is shown on the right. (**E**) Approximating a synthetic ten-dimensional dataset with known branch structure (with different colors indicating different branches), where one of the branches (blue one) extrude into higher dimensions and collapses with other branches when projected in 2D by principal component analysis (left). Middle, being applied in the space of two first principal components, ElPiGraph does not recover the branch, while it is captured when the ElPiGraph is applied in the complete ten-dimensional dataset (right). In both cases the principal tree is visualized using metro map layout[38], and a pie chart associated to each node of the graph indicates the percentage of points belonging to the different populations. The size of the pie chart indicates the number of points associated with each node.

*Construction of principal graphs to datasets containing millions of points, without pre-clustering or downsampling*

The core functionalities of ElPiGraph are based on a fast algorithm which allows the method to scale well to very large datasets. ElPiGraph.R can also take advantage of multiprocessing to further speed up the computation if necessary. When run on a single core, ElPiGraph.R is able to reconstruct principal curves and circles in few minutes even if tens of thousands points with tens of dimensions are used (Figure 3A). The construction of principal trees is significantly slower due to the combinatorial nature of the search in the structural space, but remains quite fast (Figure 3A). For comparison the R implementation of DDRTree[31] is unable to deal with large datasets in a reasonable amount of time (Figure 3C). Notably, most of the existing methods are not directly applicable to datasets having more than several thousands of points, and, hence, require data pre-clustering, downsampling or excessively drastic dimension reduction, which leads to coarse-graining of the resulting approximator structure. Multicore execution further improves the speed of the ElPiGraph.R making it able to reconstruct a tree with 60 nodes using more than one million three dimensional points in less than 3 hours with 4 cores (Figure 3B). It is worth noting that multicore principal graph construction requires additional memory management operations and may actually slow down the reconstruction for simpler problems, when compared to single core execution (Supplementary Figure 3).

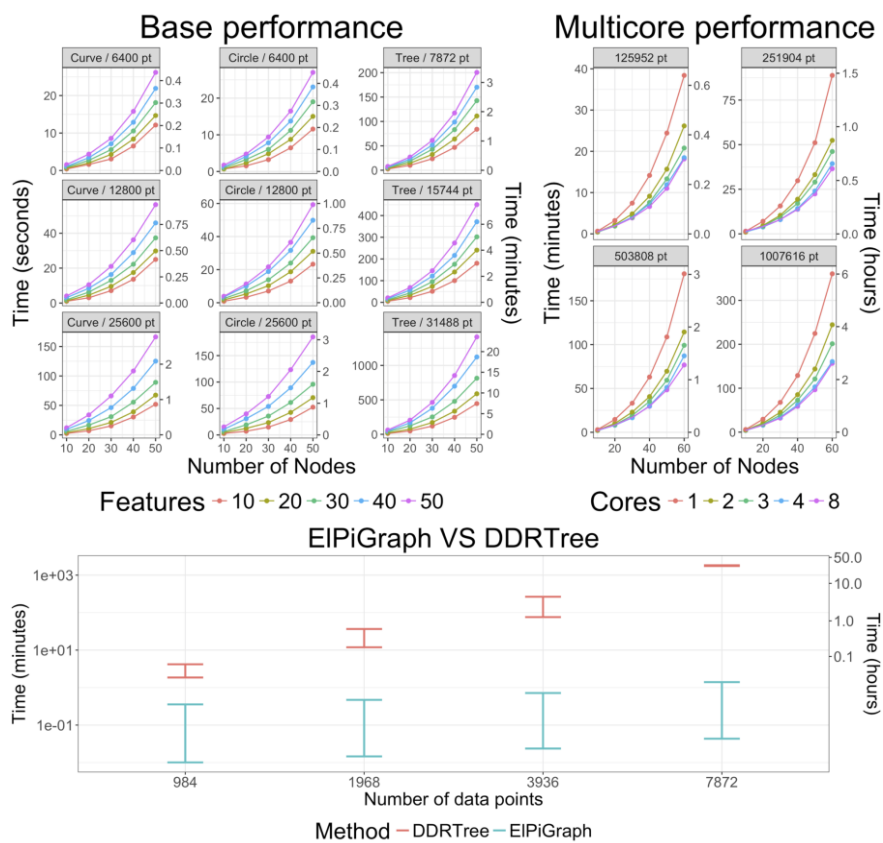

**Figure 3. Computational performance of ElPiGraph.R**. **(A)** Time required to build principal curves, circles, and trees (y axis) with a different number of nodes (x axis) using the default parameters across synthetic datasets containing different numbers of points (facets) having a different number of dimensions (color scale) without parallelization. **(B)** Time required to build principal trees (y axis) with different number of nodes (x axis) using the default parameters across synthetic datasets containing different numbers of three-dimensional points (facets) and a different number of parallel processes (color scale). **(D)** Time required to perform the DDRTree (y axis) versus the time required to build principal trees using 10-dimensional datasets with a different number of points (x axis). For DDRTree, the bar was obtained by selecting between 2 and 10 output dimensions and selecting the slower and faster execution. For ElPiGraph.R, the distribution was obtained by selecting between 10 and 50 nodes using the default parameters. All the tests were run on a CentOS 7 64bit workstation with 16GB of RAM and an Intel Xeon X5472 processor with 8 cores running at 3.00GHz.

*Inferring branching pseudo-time from single-cell RNASeq data via ElPiGraph*

Thanks to the emergence of single cell RNA Sequencing (scRNA-Seq), it is now possible to measure gene expression levels across thousands to millions of single cells. Using these data, it is then possible to look for *paths in the data* that may be associated with the level of cellular commitment w.r.t. a specific biological process and use the positions of cells along these paths (called *pseudotime*) to explore how gene expression changes as cells increase their level of commitment (Figure 4A). This kind of analysis is a powerful tool that has been used, e.g., to explore the biological changes associated with development[22], cellular differentiation[23–25], and cancer biology[26].

ElPiGraph is being used as part of STREAM[41], an integrated analysis tool that provides a set of preprocessing steps and a powerful interface to analyse scRNA-Seq or other single cell data to derive, for example, genes differentially expressed across the reconstructed paths. An extensive showcase of the power of ElPiGraph in dealing with biological data as part of the STREAM pipeline is discussed elsewhere[41]. In the present work, we will limit our analysis to a single case of scRNA-Seq dataset related to haematopoiesis[42].

Haematopoiesis is an important biological process that depends on the activity of different progenitors. To better understand this process researchers used scRNA-Seq to sequence cells across 4 different populations[42]: common myeloid progenitors (CMPs), granulo-monocyte progenitors (GMPs), megakaryocyte-erythroid progenitors (MEPs), and dendritic cells (DCs). Using the same preprocessing pipeline as STREAM, which includes selection of the most variant genes via a LOESS-based method[43] and dimensionality reduction by Modified Local Linear Embedment (MLLE)[44], we obtained a 3D projection of the original data and applied ElPiGraph with resampling (Figure 4B-E, Supplementary Figure 4).

As Figures 4B-C and Supplementary Figures 4A-B show, ElPiGraph is able to easily recapitulate the differentiation of CMPs into GMPs and MEPs on the MLLE-transformed data. A further branching point corresponding to early-to-intermediate GMPs into DCs can be observed. This differentiation trajectory is characterized by a visible level of uncertainty associated with the branching point. The emerging biological picture is compatible with DC emerging at across a specific, but relatively broad, *range of commitment* of GMPs. However, it is worth stressing that the number of DCs present in the dataset under analysis is relatively small (30 cells), and hence that the higher uncertainty level associated with the branching from GMPs to DC may be simply due to the relatively small number of DC sequenced at an early committed state.

Notably, when we used ElPiGraph on the expression matrix restricted to the most variant genes and used PCA to retain the leading 250 components, we could still discover the branching point associated with the differentiation of CMPs into GMPs and MEPs (Supplementary Figure 4F-I). However, DCs do not seem to produce an additional branching, suggesting that MLLE can be a powerful way to discover subtle differences among cell populations.

We can further explore the genetic changes associated to DC differentiation by projecting the closest trajectory, hence obtaining a pseudotime value that we can use to explore gene expression variation across branches and look for potentially interesting patterns. Figure 4D shows the dynamics of set of notable genes, with more genes selected to their high variance, large mutual information when looking at the pseudotime ordering, and significant differences across the diverging branches reported by Supplementary Figure 4C-E. Note that in all of the plots, the pseudotime with value 0.5 corresponds to the DC-associated branching point and the vertical grey area indicate a 95% confidence interval computed using the position of branching points of the resampled principal graphs. The confidence interval provides a good indication of the uncertainty associated with the determination of the branching points and provides, to a first approximation, an indication of the pseudotime range when the transcriptional programs of the two cell populations start to diverge.

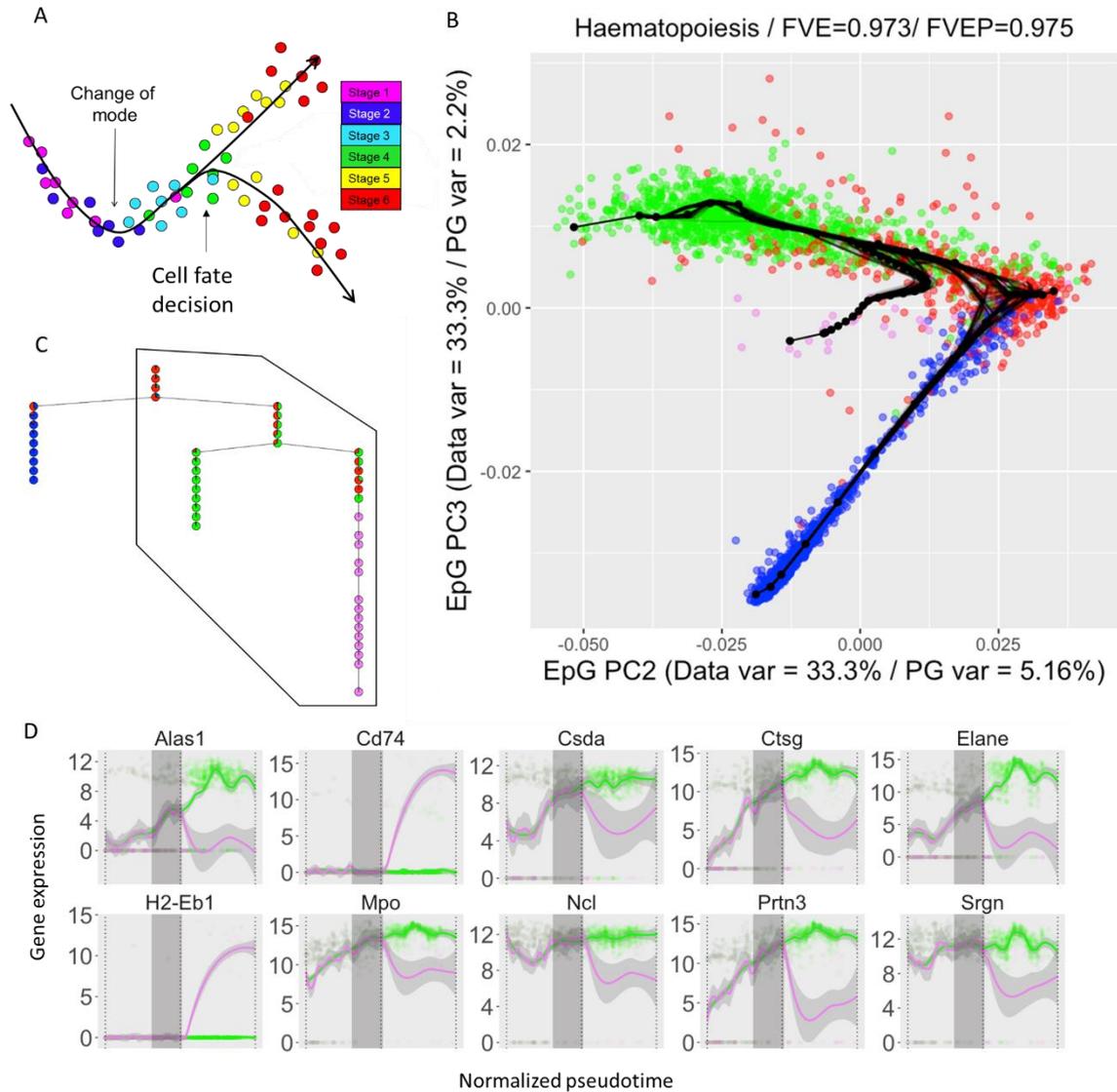

**Figure 4. ElPiGraph is able to quantify biological pseudotime from single cell data. (A)** Diagrammatic representation of the concept behind biological pseudotime in an arbitrary 2D space associated with gene expression: as cells progress from Stage 1 they differentiate (Stage 2 and 3) and branch (Stage 4) into two different subpopulations (Stage 5 and 6). Local distances between the cells indicate genetic similarity. Note how embedding a tree into the data allow recovering genetic changes associated with cell progression into the two differentiated states. **(B)** Application of ElPiGraph to scRNA-Seq data[42]. Each point indicates a cell and is color-coded using the cellular phenotype inferred by the source paper (CMP in red, DC in violet, GMP in green, and MEP in blue). One hundred bootstrapped trees are represented (in black), along with the a tress fitted on all the data (black nodes and edges). The 2D projection has been obtained by selecting the ElPiGraph principal components 2 and 3. The percentage of variance explained by the projections on the two dimensions of the data (Data var) and nodes of the tree (PG var) are reported along with the fraction of variance of the original data explained by the projection on the nodes (FVE) and on the edges (FVEP). **(C)** Diagrammatic representation of the distribution of cells across the branches of the tree reconstructed by ElPiGraph with the same color scheme used in panel B. Pie charts indicates the distribution of populations associated with each node. The black polygon highlights the subtree used to study gene pseudotime. **(D)** Variation of gene expression along the path from the root of the tree (at the top of panel C) to the branch corresponding to DC commitment and GMP commitment. Points in the background represent cells and are colored to show the path they belong to. The expression profiles have been fitted by a LOESS smoother, colored according to the majority cell type in the branch, with a 95% confidence interval (in grey). The vertical grey area represents a 95% confidence interval obtained by projecting the relevant branching points of the resampled tree showed in panel B on the path of the principal graph.

*Approximating the complex structure of development or differentiation from single cell RNASeq data*

Recently, several large-scale experiments designed to derive single-cell snapshots of a developing embyo[1,2] or differentiating cells in an adult organism[3] have been produced. Standard force-directed layout algorithms applied to *k*NN graphs connecting single cells in a reduced-dimension transcriptomic space was capable of producing informative representations of these large datasets[45]. However, such representations can be characterised by complex point distributions, with areas of varying density, varying local dimensionality and excluded regions. Hence, it may be helpful to derive *data cloud skeletons*, which would simplify the comprehension and study of these distributions.

To this end, we used scRNA-Seq obtained from stage 22 Xenopus embryos[1] to derive a 3D force directed layout projection (Figure 5A). Given the complexity of the data, we decided to employ an advanced multistep analytical pipeline, based on ElPiGraph. As a first step, we fitted a total 1280 principal trees with ten different trimming radiuses (to account for the differences in data density across the data space). This resulted in 10 *bootstrapped principal forests* (Figure 5B). Then, for each principal forest, we built a *consensus graphs*, which summarizes their features (Figure 5C). A final consensus graph was then built by combining the previously obtained ones (Figure 5D). This graph was then filtered and extended to better capture the data distribution (Figure 5E). From this analysis, clear non-trivial structures emergence: linear paths, interconnected closed loops, and branching points can be clearly observed.

The different branches (defined as linear path between nodes of degree different from two), display a statistically significant (Chi-Squared test p-val $< 5 \cdot 10^{-4}$) associations with previously defined populations[1] (Figure 5F). Using the principal graph obtained, it is also possible to obtain a pseudotime that can be used to explore how different genes vary across the different branches (Figure 5G, Supplementary Figure 6). Notably, our approach is able to identify structured transcriptomic variation in a group of cells previously identified as "outliers" (Figure 5G, top panel). Furthermore, note how our approach identify a loop in the part of the graph associated with the neural tube (Figure 5F, top left), suggesting the presence of complex *diverging-converging* cellular dynamics.

The same analysis pipeline was used to explore the transcriptome of the whole-organism data obtained from planarians[3]. As in the case of xenopus development data, a complex structure (Figure 5H) displaying a statistically significant association (Chi-Squared test p-val $< 5 \cdot 10^{-4}$) with previously defined cell types emerges (Figure 5I). As before, such structure can be used to identify how gene dynamics changes as cells commit to a specific cell type (Figure 5I, Supplementary Figure 8).

*Approximating the large scale-structure of the Universe via ElPiGraph*

Astronomy is a classically data rich discipline, with the data collected by the Danish astronomer Tycho Brahe arguably being one of the first documented examples of scientific Big Data. Nowadays, curated astronomical data catalogues containing extensive information on many features of large and small celestial objects are available to the scientific community to explore the features of the Universe. In particular, the positions of galaxies in the 3D space of galaxy redshift velocities are likely to provide important information on initial conformation of the Universe[46].

To explore the potential of ElPGraph in this domain, we obtained the V8k catalogue, which contains the supergalactic coordinates in the redshift space of 30,124 galaxies with velocities smaller than 8000 km/s[46]. Even by visual inspection, it is quite easy to identify different large-scale *filaments* present in the data. The complex distribution of the data clearly limits the application of simple manifold learning techniques that assume a predetermined topology.

As in the case of force-directed layouts, we used a multistep approach. Initially, we used 100 resampling trees, with initial starting points randomly placed in the densest region of the data. Then, we removed the points captured by at least 20 of the trees and repeated the procedure until the majority of the points were associated to at least 20 trees. This lead to a bootstrapped principal forest (Figure 6A), which we then used to construct a set of consensus graphs (Figure 6B). This example further shows the ability of ElPiGraph to extract structural information even when the expected topology of the data under consideration is unknown or too complex to be described by simple grammars.

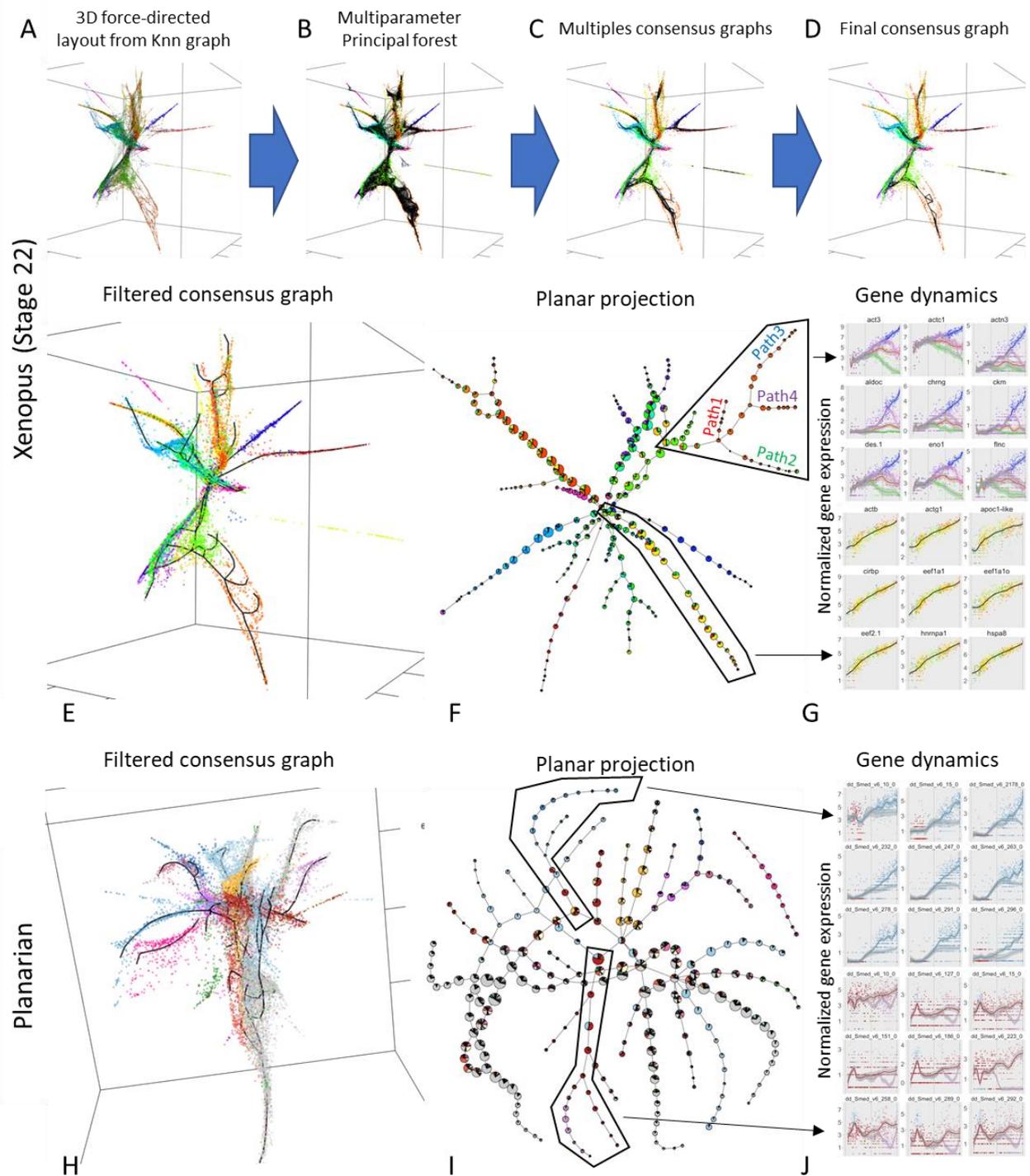

**Figure 5. ElPiGraph is able to approximate complex datasets describing developing embryos (xenopus) or adult organisms (planarian).** (**A**) A kNN graph constructed using the gene expression of 7936 cells of Stage 22 Xenopus embryo has been projected on a 3D space using a force directed layout. The color in this and the related panels indicate the population assigned to the cells by the source publication. (**B**) The coordinates of the points in panel A have been used to fit 1280 principal trees with different parameters, hence obtaining a principal forest. (**C**) The principal forest shown in panel B has been used to produce 10 consensus graphs (one for each parameter set). (**D**) A final consensus graph has been produced using the consensus graphs shown in panel C. (**E**) A final principal graph has been obtained by applying standard grammar operations to the consensus graph shown in panel D. (**F**) The associations of the different cell types to the nodes of the consensus graph shown in panel E is reported on a plane with a pie chart for each node. Note the complexity of the graph and the predominance of different cell types in different branches, as indicated the predominance of one or few colors. (**G**) The dynamics

of notable genes had been derived by deriving a pseudotime for a branching structure (Top) and a linear structure (Bottom) present in the principal graph of panel E (see black polygons). Each point represent the gene expression of a cell and their color indicate either their associated path (top) of the cell type (bottom). The gene expression profiles have been fitted with a LOESS smoother which include a 95% confidence interval. In the top panel the smoother has been colored to highlight the different paths, with the color indicated in the text of panel F. **(H-J)** The same approach described by panels A-G has been used to study the single-cell transcriptome of planarians. In panel J the color of the smoother indicates the predominant cell type on the path. The color codes used in panels are described by Supplementary Figure 5 and 7. Interactive versions of key panels are available at https://sysbio-curie.github.io/elpigraph/.

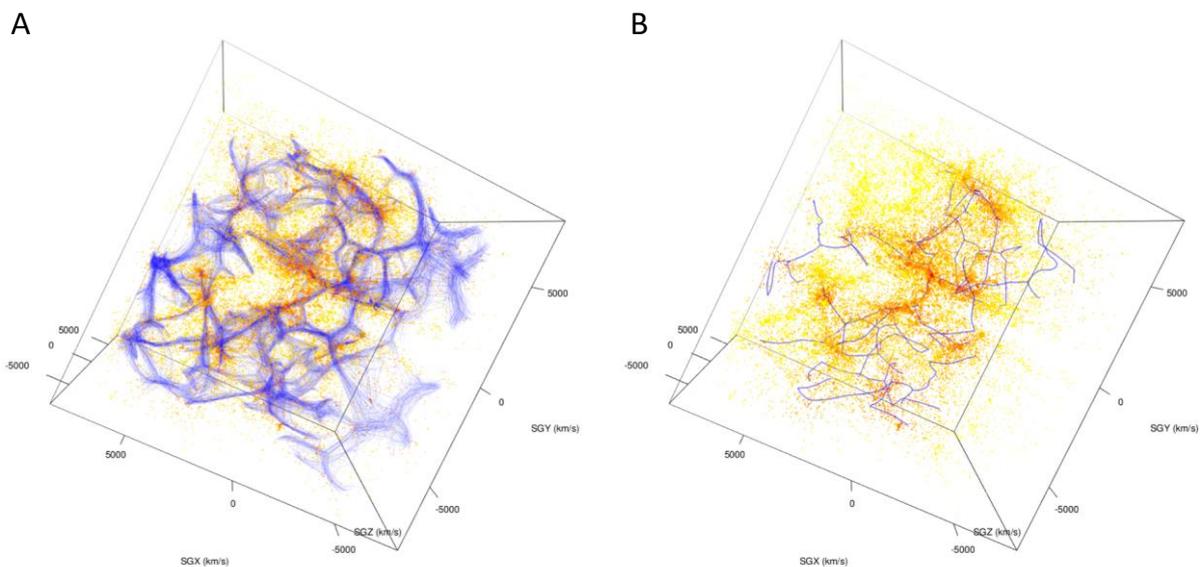

**Figure 6. ElPiGraph is able to explore complex astronomical data. (A)** A bootstrapped principal forest (blue lines) constructed on the V8k catalogue. **(B)** The disconnected consensus graph (blue lines) obtained from the bootstrapped principal forest presented by panel A. In both panels, points represent galaxies the axes indicate cartesian components of the velocity in the redshift space. The color of the points shows the distance from the closest principal graph, with stronger shades or red indicating shorter distances. The interactive versions of the panels are available at https://sysbio-curie.github.io/elpigraph/.

**Discussion (771 words)**

Bigger, and more complex, data are becoming more and more common across many scientific disciplines and being able to extract the relevant features of these data is an important step in deriving and testing scientific hypotheses. ElPiGraph represents a flexible approach for approximating a cloud of data points in a multidimensional space by reconstructing a one-dimensional continuum passing through the *middle of the data point cloud*. This approach is similar to principal curve fitting, but allows significantly more complex topologies with, for example, branching points, self-intersections and closed paths. ElPiGraph is specifically designed to be robust with respect to the noise in the data and scales up to millions of multidimensional points. These features, together with an explicit control of the structural graph complexity, make ElPiGraph applicable in many scientific domains, from data analysis in molecular biology to image analysis, and to the analysis of complex astronomical data.

Determining the topology of a high-dimensional dataset approximator without strong a priori constraints remains a challenging problem due to the combinatorial complexity of this task. In the case of highly noisy data sample or complex (e.g., self-intersecting) data approximator structure, the problem can be sometimes even ill-posed.

However, even in these circumstances, ElPiGraph is able to extract *a data structure skeleton* which can provide a useful approximation to be used for data visualization and clustering.

By specifying in advance a space of suitable graph structures and by penalizing excessively complex topologies, graph grammars allow the efficient exploration of a relevant subspace of the combinatorial graph structure space. ElPiGraph exploits more exhaustively candidate graph structures while searching the best structure when compared to competing methods which usually rely on heuristic approaches in order to define one graph structure for a given cloud of data points.

One of the key problem in finding an intrinsic data manifold topology is the presence of multiple local minima in the optimization criteria. Hence it is very important to define problem-specific strategies that are helpful to guess an initial graph structure. Under most circumstances the structure exploration procedure of ElPiGraph is capable to improve the initial guess graph structure despite the fact that this guess may be too complex, or too coarse-grained. This is achieved by applying an optimal sequence of graph structure simplification, or complexification, via a set of predefined graph rewriting rules.

Another important tool available in ElPiGraph to avoid spurious atypical local minima is its ability to apply bootstrapping based on resampling. ElPiGraph is able to employ this approach even for rather large datasets, due to its highly optimized core algorithm, which allows approximating the data by an ensemble of principal graphs and, when necessary, by constructing a consensus principal graph, as we have shown.

It is worth stressing that efficient data approximation algorithms do not overcome the need for careful data pre-processing, selection of the most informative features, and filtering of potential artefacts present in the data; since the resulting data manifold topology crucially depends on these steps. The application of ElPiGraph to scRNA-Seq data describing haematopoiesis clearly exemplifies this aspect, as the use of MLLE was able to highlight important features of the data that were otherwise hard to distinguish.

The methodological approach employed by elastic principal graphs is not limited to reconstructing intrinsically one-dimensional data approximators. Similar to self-organizing maps, principal graphs organized into regular grids can effectively approximate data by manifolds of relatively low intrinsic dimensions (up to four dimensions in practice due to the exponential increase in the number of nodes). Previously such approach was implemented in the method of elastic maps[11,18], which requires introducing non-primitive elastic graphs characterized by several selections of subgraphs (stars) in the elastic energy definition. The method of elastic maps has been successfully applied for non-linear data visualization, within multiple domains[12,47–50]. Conceptually, it remains an interesting research direction to explore the application of elastic principal graph framework to reconstructing intrinsic data manifolds characterized by varying intrinsic dimension.

Altogether, ElPiGraph enables construction of flexible and robust approximators of large datasets characterized by high topological and structural complexity. Furthermore, ElPiGraph is does not rely on a specific feature selection or dimensionality reduction techniques, making is easily integrable into different pipelines. Ensembles of principal trees allows more robust construction of complex data approximators compared to any method constructing a single tree-like approximation. Ensemble-based approach provides confidence estimations on the inferred non-trivial data features (such as branching or excluded regions) and allows constructing consensus principal graph which topology might be more complex compared to a simple tree. All in all, this indicates that ElPiGraph could be an invaluable tool for the increasingly complex data landscape of the scientific literature.

**Methods (1767 words)**

*Elastic principal graphs: basic definitions*

Let $G$ be a simple undirected graph with a set of nodes $V$ and a set of edges $E$. Let $|V|$ denote the number of nodes of the graph, and $|E|$ denote the number of edges. Let a *k-star* in a graph $G$ be a subgraph with $k + 1$ nodes $v_{0,1,...,k} \in V$ and $k$ edges $\{(v_0, v_i)/i = 1, .., k\}$. Let $E^{(i)}(0), E^{(i)}(1)$ denote the two nodes of a graph edge $E^{(i)}$, and $S^{(j)}(0), ... , S^{(j)}(k)$ denote nodes of a $k$-star $S^{(j)}$ (where $S^{(j)}(0)$ is the central node, to which all other nodes are connected). Let $\deg(v_i)$ denote a function returning the order $k$ of the star with the central node $v_i$ and 1 if $v_i$ is a terminal node. Let $\phi: V \rightarrow \mathbf{R}^m$ be a map which describes an embedding of the graph into a multidimensional space by mapping a node of the graph to a point in the data space. For any $k$-star of the graph $G$ we call its embedding *harmonic* if the embedding of its central node coincides with the mean of the leaf embeddings, i.e. $\phi(central\ node) = \frac{1}{k}\sum_{i=1...k} \phi(i)$.

We define an *elastic graph* as a graph with a selection of families of *k*-stars $S_m$ and for which all $E^{(i)} \in E$ and $S_m^{(j)} \in S_k$ have associated elasticity moduli $\lambda_i > 0$ and $\mu_j > 0$. Furthermore, a *primitive elastic graph* is defined as an elastic graph in which every non-leaf node (i.e., with at least two neighbors) is associated with a *k*-star formed by *all* the neighbors of the node. All *k*-stars in the primitive elastic graph are in the selection, i.e. the $S_k$ sets are completely determined by the graph structure. Non-primitive elastic graphs are not considered here, but they can be used, for example, for constructing 2D and 3D elastic principal manifolds, where a node in the graph can be a center of two 2-stars, in a rectangular grid[12].

For brevity, we also define an *elastic principal tree* as an acyclic primitive elastic principal graph.

*Elastic energy functional*

The *elastic energy of the graph embedment* is defined as a sum of squared edge lengths (weighted by the elasticity moduli $\lambda_i$) and the sum of squared *deviations from harmonicity* for each star (weighted by the $\mu_j$):

$$U^\phi(G) = U_E^\phi(G) + U_R^\phi(G),$$

$$U_E^\phi(G) = \sum_{E^{(i)}} \left[\lambda_i + \alpha \left(\max\left(2, \deg\left(E^{(i)}(0)\right), \deg\left(E^{(i)}(1)\right)\right) - 2\right)\right] \left(\phi(E^{(i)}(0)) - \phi(E^{(i)}(1))\right)^2,$$

$$U_R^\phi(G) = \sum_{S^{(j)}} \mu_j \left(\phi(S^{(j)}(0)) - \frac{1}{\deg(S^{(j)}(0))} \sum_{i=1}^{\deg(S^{(j)}(0))} \phi(S^{(j)}(i))\right)^2.$$

The term describing the deviation from harmonicity in the case of 2-star is a simple surrogate for minimizing the local curvature. In the case of *k*-stars with *k*>2 it can be considered as a generalization of local curvature defined for a branching point[12].

*Optimization functional for fitting a graph to data*

Assume that we have defined a partitioning *K* of all data points such that $K(i) = \arg\min_{j=1...|V|} \|X_i - \phi(V_j)\|$ is an index of a node in the graph which is the closest to the *i*th data point among all graph nodes. We define the optimization functional for fitting a graph to data as a sum of the approximation error and the elastic energy of graph embedment:

$$U^\phi(X, G) = \frac{1}{\sum w_i} \sum_{j=1}^{|V|} \sum_{K(i)=j} w_i \cdot \min\left(\|X_i - \phi(V_j)\|^2, R_0^2\right) + U^\phi(G),$$

where $w_i$ is a weight of the data point *i* (can be equal one for all points), $|V|$ is the number of nodes, $\|..\|$ is the usual Euclidean norm and $R_0$ is a trimming radius that can be used to limit the effect of *points distant from the current configuration of the graph*[51].

The objective of the basic optimization algorithm is to find a map $\phi: V \to \mathbf{R}^m$ such that $U^\phi(X,G) \to \min$ over all possible elastic graph *G* embeddings into $\mathbf{R}^m$. In practice, we are looking for a local minimum of $U^\phi(X,G)$ by applying the standard splitting-type algorithm, which is described by the pseudo-code provided in Supplementary Information text. The essence of the algorithm (similar to the simplest *k*-means clustering) is to alternate between 1) computing the partitioning *K* given the current estimation of the map $\phi$ and 2) computing new map $\phi$ provided the partitioning *K*. The first task is a standard neighbor search problem, while the second task is solving a systems of linear equations of size $|V|$.

*Graph grammar approach for determining the optimal graph topology*

In contrast to the majority of existing state-of-the-art methods, ElPiGraph exploits a topological graph grammar-based approach to find an optimal structure, by systematic application of some pre-defined set of graph grammar operations. This allows the algorithm to explore many structures in the space of all possible graph structures by using a gradient descent-like algorithm (Figure 1A). In the simplest case, two graph grammar operations "bisect an edge" and "add node to a node" allows exploring the space of trees (Supplementary Figure 2E); in this case

ElPiGraph allows constructing principal trees[39]. For a detailed definition of graph rewriting rules, see Supplementary Information text.

*Principal graph ensembles*

Resampling techniques are often used to determine the significance of analyses applied to complex datasets[52,53]. Thanks to its performance, ElPiGraph can be also used in conjunction with resampling to explore the robustness of the reconstruct principal graph. Figure 1C shows a simple example of data resampling applied to the reconstruction of a principal tree. Resampling produces a *principal graph ensemble*, i.e., a set of *k* principal graphs produced by random sampling of *p*% of data points and applying ElPiGraph *k* times (in Figure 1C, *k* = 100 and *p* = 90%). From this example, it is possible to judge how a different level of uncertainty is associated with different parts of the tree and to explore the uncertainty associated with a branching point.

*Construction of the consensus graph*

In complex examples, a principal graph ensemble can be used to infer a *consensus principal graph*, i.e. a principal graph obtained by recapitulating the information from the graph ensemble in such a way to discover emergent complex topological features of the data. For example, a consensus principal graph constructed by combining an ensemble of trees may contain loops, which are not present in any of the constructed principal trees (Figure 5 and Figure 6).

To integrate multiple principal graphs $G_{i..n}$ into a consensus principal graph $C$ with a given number of nodes $M$ is a multi-step process that can be implemented using different approaches. Briefly, all the nodes of the original principal graph $v^{G_1}_{1,\ldots,|G_1|}, v^{G_2}_{1,\ldots,|G_2|}, \ldots, v^{G_n}_{1,\ldots,|G_n|}$ are associated by an assignment operator $f$ to $M$ clusters so that $f(v) = k$ ($k = 1..M$). This clustering is performed on the positions of the nodes, such that nodes assigned to the same cluster will be in the same area of the space. Each cluster $i$ will be associated to a node $v^C_i$ of C. The coordinate of $v^C_i$ is then obtained by computing the centroid of nodes of the cluster $f^{-1}(i)$. An edge is placed between two nodes $v^C_i$ and $v^C_j$ if the number of edges between the set of nodes $f^{-1}(i)$ and the set of nodes $f^{-1}(j)$ is larger than a given threshold. The number of edges between the two sets of nodes is then used to assign a weight to the edge.

The function used to construct the consensus graph in ElPiGraph.R exploits the *k*-means algorithm to cluster the nodes and support additional features to filter the graph. Specifically, it is possible 1) to ignore nodes belonging to the original principal graphs with a low local density of nodes and hence potentially connected with *outliers principal graphs*, 2) to filter the consensus graph so that unconnected nodes are removed, and to 3) filter edges of the consensus graph that are shorter, or longer, than given thresholds.

*Leaf extension and Branch filtering*

The ElPiGraph algorithm is designed to place nodes at the centre of set of points. As a consequence of this, the leaf nodes of a principal graph might not capture the extreme points of the data distribution. This is not ideal when pseudotime analysis is performed. Hence, ElPiGraph.R includes a function that can be used to extend the leaf nodes by extrapolation. Given a leaf node, the basic strategy consists in selecting the points associated with the leaf node, but not projected on any edge. Then these points are used to produce a linear interpolation originating from the leaf node. This strategy can be implemented by using a different approach that are described in the help of ElPiGraph.R.

Under certain circumstances, it may be necessary to simplify a principal graph by removing edges or nodes that capture only a small subset of points. For this purpose, ElPiGraph.R includes a function that remove edges on which only a limited number of points are projected. The nodes of the graph are then removed or adjusted to account for the edge removal if necessary.

*Code availability*

The ElPiGraph method is implemented in several programming languages:

- R from https://github.com/sysbio-curie/ElPiGraph.R
- MATLAB from https://github.com/sysbio-curie/ElPiGraph.M

- Python from https://github.com/sysbio-curie/ElPiGraph.P

A Java implementation of ElPiGraph is available as part of VDAOEngine library (https://github.com/auranic/VDAOEngine/) for high-dimensional data analysis developed by Andrei Zinovyev. The Java implementation of ElPiGraph is not actively developed and the implementation of ElPiGraph in Java does not scale as good as other implementations.

A Scala implementation is also available from https://github.com/mraad/elastic-graph.

All the implementations contain the core algorithm as described in this paper. However, the different implementations differ in the set of functionalities improving the core algorithm such as robust local version of the algorithm, boosting up the algorithm performance by local optimization of candidate graphs, using the resulting graphs in various applications, or multicore implementation.

The code used to perform the analysis, to generate the figures, and interactive versions of some of the figures are available at: https://sysbio-curie.github.io/elpigraph/

*Processing of single cell RNA-Seq data*

All the scRNA-Seq datasets were preprocessed using a pseudocount transformation of the original row count. The LOESS-based feature selection was performed as described in STREAM pipeline. Hematopoietic data were transformed using PCA and selecting the first 250 dimensions or MLLE.

The force directed layouts where constructed by first projecting the data onto the first 20 principal components. Then a distance matrix was computed and a kNN graph (k = 5) was constructed. Then 1000 iterations of the 3D Kamada-Kawai algorithm were applied using the distances associated to the edges previously computed as weights. The output of the Kamada-Kawai algorithm was used as the initial configuration of the Fruchterman-Reingold algorithm, which was run for 2000 iteration with weights equal to 500 multiplied by the inverse of the distances associated to the edges. Graph manipulation and layout derivation were performed using the R version of the igraph library (igraph.org).

Pseudotime was computed by projecting the cells onto the closest edge and normalized in such a way that the start of the pseudotime is at 0, the end for each branch at 1, and that the branching points are equally spaced between 0 and 1.

**Author contributions**

LA, EM, AG, AZ provided original concepts and developed the methodology. LA, EM, AM, LF, AZ implemented the algorithm in several programming languages. LA, HC preprocessed the data. LA, AZ analyzed the data. LA, AZ drafted the manuscript. EB, AZ, AG supervised the project. All the authors have read and edited the manuscript.

**Competing interests**

The authors declare no competing interests.

**Acknowledgements**

This work is supported by Ministry of Education and Science of Russia (Project No. 14.Y26.31.0022), by ITMO Cancer SysBio program (MOSAIC) and INCa PLBIO program (CALYS, INCA_11692). This project has been made possible in part by grant number 2018-182734 from the Chan Zuckerberg Initiative DAF, an advised fund of Silicon Valley Community Foundation.
**References (53)**

1. Briggs, J. *et al.* The dynamics of gene expression in vertebrate embryogenesis at single-cell resolution. *Science (80-. ).* **360,** eaar5780 (2018).
2. Wagner, D. *et al.* Single-cell mapping of gene expression landscapes and lineage in the zebrafish embryo. *Science (80-. ).* **360,** 981–987 (2018).
3. Plass, M. *et al.* Cell type atlas and lineage tree of a whole complex animal by single-cell transcriptomics. *Science (80-. ).* **360,** eaaq1723 (2018).

4. Gorban, A. N. & Yablonsky, G. S. Grasping Complexity. *Comput. Math. with Appl.* **65,** 1421–1426 (2013).
5. Zinovyev, A. Overcoming Complexity of Biological Systems: from Data Analysis to Mathematical Modeling. *Math. Model. Nat. Phenom.* **10,** 186–205 (2015).
6. Rayón, P. & Gromov, M. Isoperimetry of waists and concentration of maps. *Geom. Funct. Anal.* **13,** 178–215 (2003).
7. Gorban, A. N. & Tyukin, I. Y. Stochastic separation theorems. *Neural Networks* **94,** 255–259 (2017).
8. Gorban, A. N. & Tyukin, I. Y. Blessing of dimensionality: mathematical foundations of the statistical physics of data. *Philos. Trans. R. Soc. A Math. Phys. Eng. Sci.* **376,** (2018).
9. Pearson, K. 1901. On lines and planes of closest fit to systems of points in space. 559–572 (1901).
10. Kohonen, T. The self-organizing map. *Proc. IEEE* **78,** 1464–1480 (1990).
11. Gorban, A. & Zinovyev, A. Elastic principal graphs and manifolds and their practical applications. *Computing (Vienna/New York)* **75,** 359–379 (2005).
12. Gorban, A. N. & Zinovyev, A. Principal manifolds and graphs in practice: from molecular biology to dynamical systems. *Int. J. Neural Syst.* **20,** 219–232 (2010).
13. Tenenbaum, J. B., De Silva, V. & Langford, J. C. A global geometric framework for nonlinear dimensionality reduction. *Science (80-. ).* **290,** 2319–2323 (2000).
14. Roweis, S. T. & Saul, L. K. Nonlinear dimensionality reduction by locally linear embedding. *Science (80-. ).* **290,** 2323–2326 (2000).
15. Van Der Maaten, L. J. P. & Hinton, G. E. Visualizing high-dimensional data using t-sne. *J. Mach. Learn. Res.* **9,** 2579–2605 (2008).
16. Smola, A. J., Williamson, R. C., Mika, S. & Sch, B. Regularized Principal Manifolds. 214–229 (1999).
17. McInnes, L. & Healy, J. UMAP: Uniform Manifold Approximation and Projection for Dimension Reduction. *arXiv* (2018).
18. Gorban, A., Kégl, B., Wunch, D. & Zinovyev, A. Principal Manifolds for Data Visualisation and Dimension Reduction. *Lect. notes Comput. Sci. Eng.* 340 (2008). doi:10.1007/978-3-540-73750-6
19. Carlsson, G. & de Silva, V. Topological approximation by small simplicial complexes. *Preprint* 1–36 (2003).
20. Gorban, A. N. & Zinovyev, A. Y. in *Handbook of Research on Machine Learning Applications and Trends: Algorithms, Methods and Techniques* (2008). doi:10.4018/978-1-60566-766-9
21. Zinovyev, A. & Mirkes, E. Data complexity measured by principal graphs. *arXiv12125841* (2012). doi:10.1016/j.camwa.2012.12.009
22. Furlan, A. *et al.* Multipotent peripheral glial cells generate neuroendocrine cells of the adrenal medulla. *Science (80-. ).* **357,** (2017).
23. Trapnel, C. *et al.* Pseudo-temporal ordering of individual cells reveals dynamics and regulators of cell fate decisions. *Nat Biotechnol* **29,** 997–1003 (2012).
24. Athanasiadis, E. I. *et al.* Single-cell RNA-sequencing uncovers transcriptional states and fate decisions in haematopoiesis. *Nat. Commun.* **8,** 2045 (2017).
25. Velten, L. *et al.* Human haematopoietic stem cell lineage commitment is a continuous process. *Nat. Cell Biol.* **19,** 271–281 (2017).
26. Tirosh, I. *et al.* Single-cell RNA-seq supports a developmental hierarchy in human oligodendroglioma. *Nature* **539,** 309–313 (2016).
27. Cannoodt, R., Saelens, W. & Saeys, Y. Computational methods for trajectory inference from single-cell transcriptomics. *European Journal of Immunology* **46,** 2496–2506 (2016).
28. Moon, K. R. *et al.* Manifold learning-based methods for analyzing single-cell RNA-sequencing data. *Curr. Opin. Syst. Biol.* (2017). doi:10.1016/j.coisb.2017.12.008
29. Saelens, W. *et al.* A comparison of single-cell trajectory inference methods: towards more accurate and robust tools. *bioRxiv* 276907 (2018). doi:10.1101/276907
30. Drier, Y., Sheffer, M. & Domany, E. Pathway-based personalized analysis of cancer. *Proc. Natl. Acad. Sci.* **110,** 6388–6393 (2013).
31. Qiu, X. *et al.* Reversed graph embedding resolves complex single-cell trajectories. *Nat. Methods* (2017). doi:10.1038/nmeth.4402
32. Welch, J. D., Hartemink, A. J. & Prins, J. F. SLICER: Inferring branched, nonlinear cellular trajectories from single cell RNA-seq data. *Genome Biol.* **17,** (2016).
33. Setty, M. *et al.* Wishbone identifies bifurcating developmental trajectories from single-cell data. *Nat. Biotechnol.* **34,** 637–645 (2016).
34. Kégl, B. & Krzyzak, A. Piecewise linear skeletonization using principal curves. *IEEE Trans. Pattern Anal. Mach. Intell.* **24,** 59–74 (2002).
35. Wolf, F. A. *et al.* Graph abstraction reconciles clustering with trajectory inference through a topology preserving map of single cells. *bioRxiv* 208819 (2017). doi:10.1101/208819

36. Pestov, V. Indexability, concentration, and VC theory. in *Journal of Discrete Algorithms* **13,** 2–18 (2012).
37. Chávez, E., Navarro, G., Baeza-Yates, R. & Marroquín, J. L. Searching in metric spaces. *ACM Comput. Surv.* **33,** 273–321 (2001).
38. Gorban, A. N., Sumner, N. R. & Zinovyev, A. Y. Beyond the concept of manifolds: principal trees, metro maps, and elastic cubic complexes. in *Lecture Notes in Computational Science and Engineering* **58,** 219–237 (2008).
39. Gorban, A. N., Sumner, N. R. & Zinovyev, A. Y. Topological grammars for data approximation. *Appl. Math. Lett.* **20,** 382–386 (2007).
40. Babaeian, A., Bayestehtashk, A. & Bandarabadi, M. Multiple manifold clustering using curvature constrained path. *PLoS One* **10,** (2015).
41. Chen, H. *et al.* STREAM: Single-cell Trajectories Reconstruction, Exploration And Mapping of omics data. *bioRxiv* 302554 (2018).
42. Paul, F. *et al.* Transcriptional Heterogeneity and Lineage Commitment in Myeloid Progenitors. *Cell* **163,** 1663–1677 (2015).
43. Guo, G. *et al.* Serum-Based Culture Conditions Provoke Gene Expression Variability in Mouse Embryonic Stem Cells as Revealed by Single-Cell Analysis. *Cell Rep.* **14,** 956–965 (2016).
44. Zhang, Z. & Wang, J. MLLE: Modified Locally Linear Embedding Using Multiple Weights. *Adv. Neural Inf. Process. Syst.* 1593–1600 (2006).
45. Weinreb, C., Wolock, S. & Klein, A. M. SPRING: a kinetic interface for visualizing high dimensional single-cell expression data. *Bioinformatics* (2017). doi:10.1093/bioinformatics/btx792
46. Courtois, H. M., Pomarède, D., Tully, R. B., Hoffman, Y. & Courtois, D. Cosmography of the local universe. *Astron. J.* **146,** (2013).
47. Gorban, A. N. & Zinovyev, A. Visualization of data by method of elastic maps and its applications in genomics, economics and sociology. *IHES Prepr.* (2001).
48. Gorban, A. N., Zinovyev, A. Y. & Wunsch, D. C. Application of the method of elastic maps in analysis of genetic texts. *Proc. Int. Jt. Conf. Neural Networks, 2003.* **3,** (2003).
49. Failmezger, H., Jaegle, B., Schrader, A., Hülskamp, M. & Tresch, A. Semi-automated 3D Leaf Reconstruction and Analysis of Trichome Patterning from Light Microscopic Images. *PLoS Comput. Biol.* **9,** (2013).
50. Cohen, D. P. A. *et al.* Mathematical Modelling of Molecular Pathways Enabling Tumour Cell Invasion and Migration. *PLOS Comput. Biol.* **11,** e1004571 (2015).
51. Gorban, A. N., Mirkes, E. & Zinovyev, A. Y. Robust principal graphs for data approximation. *Arch. Data Sci.* **2,** 1:16 (2017).
52. Politis, D., Romano, J. & Wolf, M. *Subsampling*. (Springer, 1999).
53. Albergante, L., Blow, J. J. & Newman, T. J. Buffered Qualitative Stability explains the robustness and evolvability of transcriptional networks. *Elife* **3,** e02863 (2014).

ANNEX

**Supplementary Notes**

*Elastic matrix*

The elastic energy functional for ElPiGraph is defined as (see the main manuscript)

$$U^\phi(G) = U_E^\phi(G) + U_R^\phi(G),$$

$$U_E^\phi(G) = \sum_{E^{(i)}} \left[\lambda_i + \alpha \left(\max\left(2, \deg\left(E^{(i)}(0)\right), \deg\left(E^{(i)}(1)\right)\right) - 2\right)\right] \left(\phi(E^{(i)}(0)) - \phi(E^{(i)}(1))\right)^2,$$

$$U_R^\phi(G) = \sum_{S^{(j)}} \mu_j \left(\phi(S^{(j)}(0)) - \frac{1}{\deg(S^{(j)}(0))} \sum_{i=1}^{\deg(S^{(j)}(0))} \phi(S^{(j)}(i))\right)^2. \quad (*)$$

Note that $U_R^\phi(G)$ can be re-written as

$$U_R^\phi(G) = \sum_{S^{(j)}} \frac{\mu_j}{\deg(S^{(j)}(0))} \sum_{i=1}^{\deg(S^{(j)}(0))} \left(\phi(S^{(j)}(0)) - \phi(S^{(j)}(i))\right)^2 -$$

$$- \sum_{S^{(j)}} \frac{\mu_j}{\left(\deg(S^{(j)}(0))\right)^2} \sum_{i=1, p=1, i<>p}^{\deg(S^{(j)}(0))} \left(\phi(S^{(j)}(i)) - \phi(S^{(j)}(p))\right)^2,$$

i.e., the term $U_R^\phi(G)$ can be considered as a sum of the energy of elastic springs connecting the star centers with its neighbors (with elasticity moduli $\mu_j/\deg(S^{(j)})$) and the energy of negative (repulsive) springs connecting all non-central nodes in a star pair-wise (with negative elasticity moduli $-\mu_j/(deg(S^{(j)}))^2$). The resulting system of springs, whose energy is minimized, is shown in Figure 1A.

In simple terms, the elasticity of the principal graph contains three parts: positive springs corresponding to elasticity of graph edges (Figure 1A, Supplementary Figure 2B), negative repulsive springs describing the node repulsion to make the graph embedding function $\phi$ as smooth as possible (Supplementary Figure 2C), positive springs representing the correction term such that the smoothing would correspond to the deviation of harmonicity (Supplementary Figure 2D).

For algorithmic reasons, it is convenient to describe the structure and the elastic properties of the graph by an elastic matrix *EM(G)*. The elastic matrix is a *|V|×|V|* symmetric matrix with non-negative elements containing the edge elasticity moduli $\lambda_i$ at the intersection of rows and lines, corresponding to each pair $E^{(i)}(0)$, $E^{(i)}(1)$, and the star elasticity module $\mu_j$ in the diagonal element corresponding to $S^{(j)}(0)$. Therefore, *EM(G)* can be represented as a sum of two matrices *Λ* and *M*:

*EM(G) = Λ(G) + M(G)*,

where *Λ* is an analog of the weighted adjacency matrix for the graph *G*, with elasticity moduli playing the role of weights, and *M(G)* is a diagonal matrix having non-zero values only for nodes that are centers of starts, in which case the value indicates the elasticity modulus of the star. An example of elastic matrix is shown in Supplementary Figure 2A.

It is also convenient to represent the elastic energy in the matrix form, by transforming *EM(G)* into three auxiliary matrices Λ, $\Lambda^{star\_edges}$ and $\Lambda^{star\_leafs}$. $\Lambda^{star\_edges}$ is a weighted adjacency matrix for the edges connected to star centers, with elasticity moduli $\mu_j/k_j$, where $k_j$ is the number of edges connected to the *j*th star center. $\Lambda^{star\_leafs}$ is a weighted adjacency matrix for the negative springs (shown in green in Figure 1A), with elasticity moduli $-\mu_j/(k_j)^2$. An example of the transforming the elastic matrix *EM(G)* into three weighted adjacency matrices Λ, $\Lambda^{star\_edges}$, $\Lambda^{star\_leafs}$ is shown in Supplementary Figure A-D.

For the system of springs shown in Figure 1A, if one applies a distributed force to the nodes, then the propagation of the node perturbation will be described by a matrix which is a sum of three graph Laplacians.

$$L(G, EM(G)) = L(\Lambda) + L(\Lambda^{star\_edges}) + L(\Lambda^{star\_leafs}) \ .$$

We remind that a Laplacian matrix for a weighted adjacency matrix $A$ is computed as

$$L(A)_{ij} = \delta_{ij} \sum_k A_{kj} - A_{ij},$$

where $\delta_{ij}$ is the Kronecker delta.

*Basic optimization algorithm*

The basic optimization algorithm fits a graph of a given structure to a finite set of data vectors. We define the optimal embedding of a graph as a map $\phi: V \to \mathbf{R}^m$ that minimizes the mean squared distance between the position of graph nodes and the data points and, at the same time, minimizes the elastic energy of the graph embedment serving a penalty term for the "irregularity" of the graph embedment. The irregularity can manifest itself by stretching and non-equal distance between graph node positions (penalized by $U_E^\phi(G)$ and partly by $U_R^\phi(G)$) or by deviation from harmonicity (penalized by $U_R^\phi(G)$).

Assume that we have defined a partitioning $K$ of all data points such that $K(i) = \arg\min_{j=1...|V|} \|X_i - \phi(V_j)\|$ returns an index of a node in the graph which is the closest to the *i*th data point among all graph nodes. Then, the objective function to minimize is (see the main manuscript Method section)

$$U^\phi(X, G) = \frac{1}{\sum w_i} \sum_{j=1}^{|V|} \sum_{K(i)=j} w_i \cdot \min\left(\|X_i - \phi(V_j)\|^2, R_0^2\right) + U^\phi(G),$$

where $w_i$ is a weight of the data point $i$ (can be unity for all points), $|V|$ is the number of vertices, $\|..\|$ is the usual Euclidean distance and $R_0$ is a trimming radius that can be used to limit the effect of *points distant from the graph*[1].

The objective of the basic optimization algorithm is to a map $\phi: V \to \mathbf{R}^m$ such that that $U^\phi(X,G) \to \min$ over all possible elastic graph $G$ embeddings in $\mathbf{R}^m$. In practice, we are looking for a local minimum of $U^\phi(X,G)$ by applying the expectation-minimization type algorithm, which is described by the pseudo-code below:

---

ALGORITHM 1: BASE GRAPH OPTIMIZATION FOR A FIXED STRUCTURE OF THE ELASTIC GRAPH

1) Initialize the graph $G$, its elastic matrix $E(G)$ and the map $\phi$.
2) Compute the matrix $L(G, E(G))$
3) Partition the data by proximity to the embedded nodes of the graph (i.e., compute the mapping $K:\{X\}\to\{V\}$ of a data point $i$ to a graph node $j$)
4) Solve the following system of linear equations to determine the new map $\phi$:

$$\sum_{j=1}^{|V|} \left( \frac{\sum_{\{K(i)=j\}} w_i}{\sum_{i=1}^{|V|} w_i} \delta_{ij} + L(G, EM(G))_{ij} \right) \phi(V_j) = \frac{1}{\sum_{i=1}^{|V|} w_i} \sum_{K(i)=j} w_i X_i \ ,$$

where $\delta_{ij}$ is the Kronecker delta.

Iterate 3-4 till the map $\phi$ does not change more than $\varepsilon$ in some appropriate measure.

---

The convergence of the algorithm can be easily proven[1] since the $U^\phi(X, G)$ is a Lyapunov function with respect to the Algorithm 1. The base algorithm optimizes the graph embedment $\phi$, but contains neither a recipe for initializing the map $\phi$, nor a recipe for choosing the structure of the graph $G$. This initialization can be derived in a number of ways, starting from the structural analysis of the *kNN* graph after some dimension reduction and/or pre-clustering (e.g., computing a spanning tree) or by other heuristic approaches.

*Graph grammar-based optimization of the graph structure*

A graph-based data approximator should deal simultaneously with two inter-related aspects: determining the optimal topological structure of the graph and determining the optimal map for embedding this graph topology into the multidimensional space. An exhaustive approach would be to consider all the possible graph topologies (or topologies of a certain class, e.g., all possible trees), find the best mapping of all them into the data space, and select the best one. In practice, due to combinatorial explosion, testing all the possible graph topologies is feasible only when a small number of nodes is present in the graph, or under restrictive constraints (e.g., only trivial linear graphs, or assuming a restricted set of topologies with a pre-defined number and types of branching). Determining a globally optimal embedment of a graph with a given topology is usually challenging, because of the energy landscape complexity. This means that, in practice, one has to use an optimization approach in which both graph topology and the mapping function should be learnt simultaneously.

A graph grammar-based approach for constructing such an algorithm was suggested before[2,3]. The algorithm starts from an initial graph $G_0$ and an initial map $\phi_0(G_0)$. In the simplest case, the initial graph can be initialized by two nodes and one edge and the map can correspond to a segment of the first principal component.

Then a set of predefined grammar operations which can transform both the graph topology and the map, is applied starting from a given pair $\{G_i, \phi_i(G_i)\}$. Each grammar operation $\Psi^p$ produces a set of new graph embedments possibly taking into account the dataset $X$:

$$\{\{D^k, \phi(D^k)\}, k = 1 \ldots s\} = \Psi^p(\{G_i, \phi_i(G_i)\}, X).$$

Given a pair $\{G_i, \phi_i(G_i)\}$, a set of $r$ different graph operations $\{\Psi^1, \ldots, \Psi^r\}$ (which we call a "graph grammar"), and an energy function $\bar{U}^\phi(X, G)$, at each step of the algorithm the energetically optimal candidate graph embedment is selected as:

$$\{G_{i+1}, \phi_{i+1}(G_{i+1})\} = \mathrm{argmin}_{\{D^k, \phi(D^k)\}} \left\{ U^{\phi(D^k)}(D^k, X) : \{D^k, \phi(D^k)\} \in \bigcup_{p=1\ldots r} \Psi^p(\{G_i, \phi_i(G_i)\}, X) \right\}$$

where $\{D^k, \phi(D^k)\}$ is supposed to be optimized (fit to data) after the application of a graph grammar using ALGORITHM 1 with initialization suggested by the graph grammar application (see below).

The pseudocode for this algorithm is provided below:

---

ALGORITHM 2: GRAPH GRAMMAR BASED OPTIMIZATION OF GRAPH STRUCTURE AND EMBEDMENT

1. Initialize the current graph embedment by some graph topology and some initial map $\{G_0, \phi_0(G_0)\}$.
2. For the current graph embedment $\{G_i, \phi_i(G_i)\}$, apply all grammar operations from a grammar set $\{\Psi^1, \ldots, \Psi^r\}$, and generate a set of $s$ candidate graph embedments $\{D^k, \phi(D^k), k = 1 \ldots s\}$.
3. Further optimize each candidate graph embedment using ALGORITHM 1, and obtain a set of $s$ energy values $\{\bar{U}^{\phi(D^k)}(D^k)\}$.
4. Among all candidate graph embedments, select an embedment with the minimum energy $\{G_{i+1}, \phi_{i+1}(G_{i+1})\}$.

Repeat 2-4 until the graph contains a required number of nodes.

Note that the energy function $\overline{U}^\phi(X,G)$ used to select the optimal graph structure is not necessarily the same energy as (1-3) and can include various penalties to give less priority to certain graph configurations (such as those having excessive branching, as described below). Separating the energy functions $U^\phi(X,G)$, used for fitting a fixed graph structure to the data, and $\overline{U}^\phi(X,G)$, used to select the most favorable graph configuration, allows achieving great flexibility in defining the strategy for selecting the most optimal graph topologies.

*Simple graph grammar operations*

Two base grammar operations "bisect an edge" and "add a node to a node" are defined below.

| GRAPH GRAMMAR OPERATION "BISECT AN EDGE" | GRAPH GRAMMAR OPERATION "ADD NODE TO A NODE" |
|---|---|
| Applicable to: any edge of the graph<br>Update of the graph structure: for a given edge {A,B}, connecting nodes A and B, remove {A,B} from the graph, add a new node C, and introduce two new edges {A,C} and {C,B}.<br>Update of the elasticity matrix: the elasticity of edges {A,C} and {C,B} equals elasticity of {A,B}.<br>Update of the graph embedment: C is placed in the mean position between the embedments of A and B. | Applicable to: any node of the graph<br>Update of the graph structure: for a given node A, add a new node C, and introduce a new edge {A,C}<br>Update of the elasticity matrix:<br>*if A is a leaf node (not a star center) then*<br>   the elasticity of the edge {A,C} equals to the edge connecting A and its neighbor, the elasticity of the new star with the center in A equals to the elasticity of the star centered in the neighbor of A. If the graph contains only one edge then a predefined values is assigned.<br>*else*<br>   the elasticity of the edge {A,C} is the mean elasticity of all edges in the star with the center in A, the elasticity of a star with the center in A does not change.<br>Update of the graph embedment:<br>*if A is a leaf node (not a star center) then*<br>   C is placed at the same distance and the same direction as the edge connecting A and its neighbor,<br>*else*<br>   C is placed in the mean point of all data points for which A is the closest node |

The application of ALGORITHM 2 with a graph grammar containing only the 'bisect an edge' operation, and a graph composed by two nodes connected by a single edge as initial condition, produces an *elastic principal curve*.

The application of ALGORITHM 2 with a graph grammar containing only the 'bisect an edge' operation, and a graph composed by four nodes connected by four edges without branching, produces a *closed elastic principal curve* (called elastic principal circle, for simplicity).

The application of ALGORITHM 2 with a growing graph grammar containing both the 'bisect an edge' and the 'add a node to a node' operations and a graph composed by two nodes connected by a single edge as initial condition produces an *elastic principal tree*. In the case of a tree or other complex graphs, it is advantageous to improve the ALGORITHM 2 by providing an opportunity to 'roll back' the changes of the graph structure. This gives an opportunity to get rid of unnecessary branching or to merge split branches created in the history of graph optimization, if this is energetically justified (see Supplementary Figure 2B). This possibility can be achieved by introducing a shrinking grammar. In the case of trees, the shrinking grammar consists of two operations 'remove a leaf node' and 'shrink internal edge' (defined below). Then the graph growth can be achieved by alternating two steps of application of the growing grammar with one step of application of the shrinking grammar. In each such cycles, one node will be added to the graph.

| GRAPH GRAMMAR OPERATION "REMOVE A LEAF NODE" | GRAPH GRAMMAR OPERATION "SHRINK INTERNAL EDGE" |
|---|---|
| Applicable to: node A of the graph with deg(A)=1<br>Update of the graph structure: for a given edge {A,B}, connecting nodes A and B, remove edge {A,B} and node A from the graph<br>Update of the elasticity matrix:<br>*if B is the center of a 2-star then*<br>   put zero for the elasticity of the star for B (B becomes a leaf)<br>*else*<br>   do not change the elasticity of the star for B<br>Remove the row and column corresponding to the vertex A<br>Update of the graph embedment: all nodes besides A keep their positions. | Applicable to: any edge {A,B} such that deg(A)>1 and deg(B)>1.<br>Update of the graph structure: for a given edge {A,B}, connecting nodes A and B, remove {A,B}, reattach all edges connecting A with its neighbours to B, remove A from the graph.<br>Update of the elasticity matrix:<br>The elasticity of the new star with the center in B becomes the average elasticity of the previously existing stars with the centers in A and B<br>Remove the row and column corresponding to the vertex A<br>Update of the graph embedment: B is placed in the mean position between A and B embedments. |

Note that, when applicable, the above operations, can be restricted so that they are applied only to nodes with a certain ranges of degrees. This can be helpful, for example, if the applied grammar is designed to better explore the vicinity of branching points or leaf nodes.

*Robust local construction of elastic principal graphs*

In the form of the ElPiGraph optimization criterion $U^\phi(X, G)$, data points located farther than $R_0$ (a parameter called "trimming radius") from any graph node position do not contribute, for a given data point partitioning, to the optimization equation in Algorithm 1. However, these data points might appear at a distance smaller than $R_0$ at the next algorithm iteration: therefore, it is not equivalent to permanently pre-filtering "outliers". The approach is similar to the "data-driven" trimmed k-means clustering[4].

The robust version of the algorithm can tolerate significant amount of uniformly distributed background noise (see Figure 2C) and even deal with self-intersecting data distributions (Figure 2D), if the trimming radius is properly chosen. ElPiGraph includes a function for estimating the coarse-grained radius of the data based on local analysis of density distribution, which can be used as a good initial guess for the $R_0$ value. An alternative initial guess for the the trimming radius can be obtained by taking the median of distribution of the distances between all pairs of points in the data.

In case of existence of several well-separable clusters in the data, with the distance between them larger than $R_0$, the robust version of ElPiGraph can approximate the principal graph only for one of them, completely disregarding the rest of the data. In this case, the approximated part of the data can be removed and the robust ElPiGraph can be re-applied. For example, this procedure will construct a second (third, fourth, etc.) principal tree. Such an approach will approximate the data by disconnected "principal forest".

Alternative ways of constructing robust principal graphs include using piece-wise quadratic subquadratic error functions (PQSQ potentials)[5], which uses computationally efficient non-quadratic error functions for approximating a dataset. These two approaches will be implemented in the future versions of ElPiGraph.

*Explicit control of principal graph complexity*

In many circumstances, it may be important to control the *level of branching* of the data approximator. For example, this is the case if some *a priori* information is available on the data structure or on the level of noise present. To deal with this situation ElPiGraph can be used with the *tuning parameter* α, which allows penalizing the appearance of complex branching points. In particular, if α=0 branching is not penalized, while larger values progressively penalize branching, with α≈1 resulting in branching being completely forbidden (see definition of the elastic energy functions). Supplementary Figure 1 illustrates, using the standard iris and a synthetic dataset, how changing this parameter eliminates non-essential branches of a fitted principal tree, up to prohibiting them and simplifying the principal tree structure to a simple principal curve. Without this penalty term, extensive branches can appear in the regions of data distributions which can be characterized by a "thick turn", i.e., when the increased local curvature of the intrinsic underlying manifold leads to increased local variance of the dataset (Supplementary Figure 1).

In case of presence of excessive branching, it is recommended to set up the first branching control parameter α to a small value (e.g., 0.01). Changing the value of α from 0 to a large value (e.g., 1) allows gradual change from the absence of excessive branching penalty to effective interdiction of branching (thus, constructing a principal curve instead of a tree as a result, see Supplementary Figure 1).

Theoretical considerations suggests that the most effective way to include this penalty is within the elastic stretching part $U_E^\phi(G)$ of the elastic energy. To illustrate this aspect, let us consider graph structures each having 11 nodes and 10 edges of equal unity length. Then, for example, the 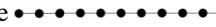 graph is characterized by a $U_E^\phi(G) = 10\lambda$ contribution to the elastic energy. The graph with one star 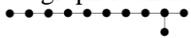 will be characterized by a $U_E^\phi(G) = 10\lambda + 3\alpha$ penalty, 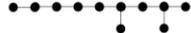 by $U_E^\phi(G) = 10\lambda + 6\alpha$, and 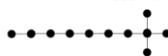 by $U_E^\phi(G) = 10\lambda + 8\alpha$.

*Strategies for graph initialization*

The construction of elastic principal graphs in ElPiGraph can be organized either by graph growth (similar to divisive clustering) or by shrinking the graph (similar to agglomerative clustering) or by exploring possible graph structures having the same number of nodes. These different strategies can be achieved by specifying the appropriate graph grammars in the parameter set. The the initial graph structure can have a strong influence on the final graph mapping to the data space and its structure.

The default setting of ElPiGraph initializes the graph with the simplest graph containing two nodes oriented along the first principal component: this initialization is able to correctly fit data topology in relatively simple cases. Other initializations are advised in the case of more complex distributions: for example, applying pre-clustering and computing (once) the minimal spanning tree between cluster centroids can be used for the initialization of the principal tree (e.g., a similar approach in used by the STREAM pipeline[6]).

When using a finite trimming radius $R_0$ value, the graph growing can be initialized by 1) a rough estimation of local data density in a limited number of data points and 2) placing two nodes, one into the data point characterized by the highest local density and another node is placed into the data point closest to the first one (but not coinciding).

*Principal forest: a way to approximate discontinuous data distributions*

As it was shown in the context of the Travel Maze problem, ElPiGraph is capable of dealing with disconnected graphs. This feature is helpful if the data to be approximated are composed of separate clusters. Under these circumstances, the local version of principal graphs should be used (by setting the trimming radius to a finite value) with appropriate initial conditions. Then, the data points approximated by the graph are removed from the data and the construction of principal graphs is repeated until no points remain associated to a graph. Note that resampling can also be used at any step if necessary.

1. Gorban, A. N., Mirkes, E. & Zinovyev, A. Y. Robust principal graphs for data approximation. *Arch. Data Sci.* **2,** 1:16 (2017).

2. Gorban, A. N. & Zinovyev, A. Y. in *Handbook of Research on Machine Learning Applications and Trends: Algorithms, Methods and Techniques* (2008). doi:10.4018/978-1-60566-766-9

3. Gorban, A. N., Sumner, N. R. & Zinovyev, A. Y. Topological grammars for data approximation. *Appl. Math. Lett.* **20,** 382–386 (2007).

4. Cuesta-Albertos, J. A., Gordaliza, A. & Matrán, C. Trimmed k-means: An attempt to robustify quantizers. *Ann. Stat.* **25,** 553–576 (1997).

5. Gorban, A. N., Mirkes, E. M. & Zinovyev, A. Piece-wise quadratic approximations of arbitrary error functions for fast and robust machine learning. *Neural Networks* **84,** 28–38 (2016).

6. Chen, H. *et al.* STREAM: Single-cell Trajectories Reconstruction, Exploration And Mapping of omics data. *bioRxiv* 302554 (2018).

**Supplementary Figures**

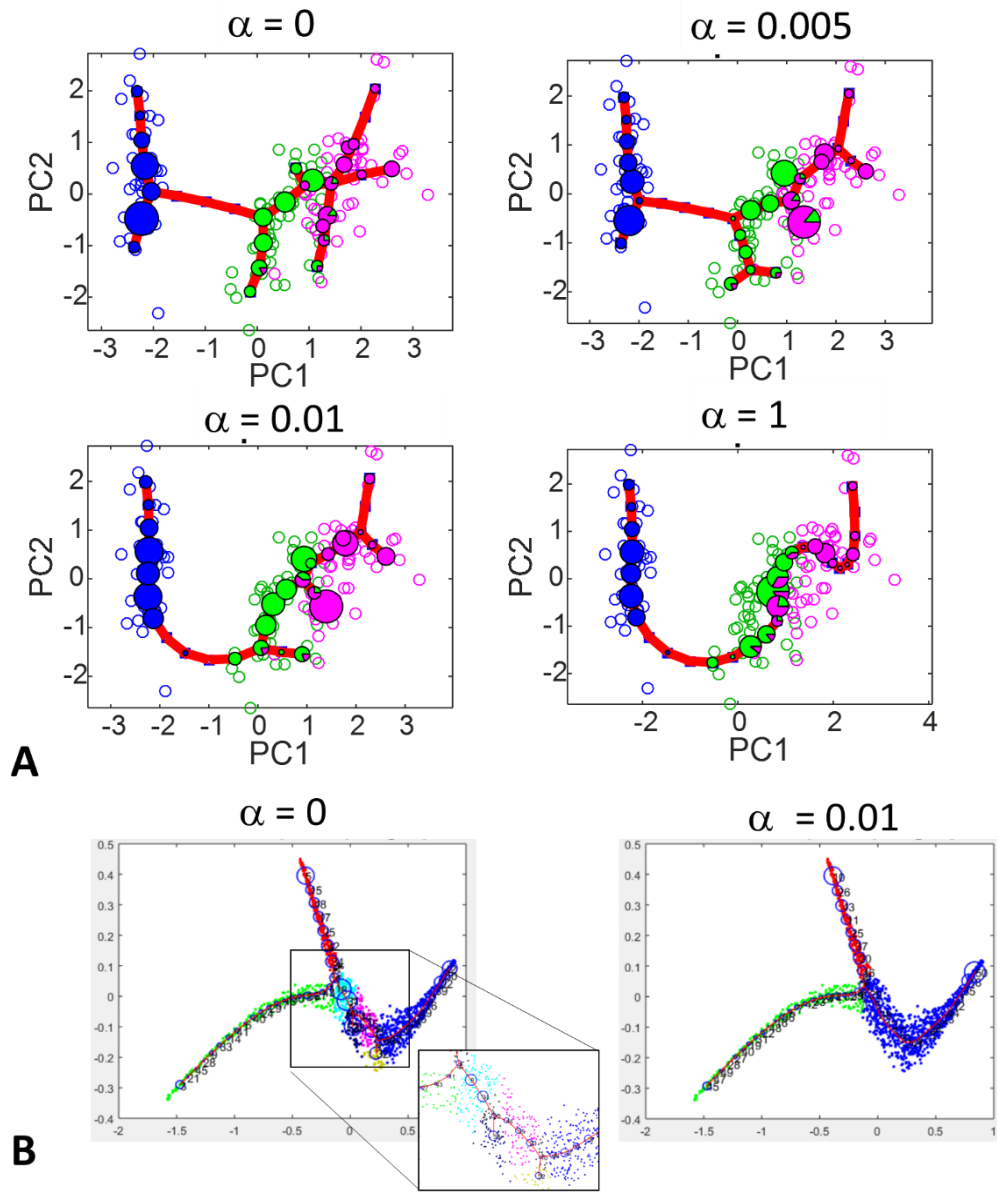

**Supplementary Figure 1. Explicit control for topological complexity in ElPiGraph, using the α parameter.** (A) Iris dataset, approximated by ElPiGraph with default parameters, using increasing values of α. (B) Synthetic dataset characterized by a "thick turn" pattern (when the local variance of the dataset increases in the region characterized by the largest curvature of the principal curve). Using an explicit control for topological complexity, it is possible to suppress the small branches while retaining the major one. Small fictitious branches appear here due to the effective increase of the local data dimension, which does not change the data topology.

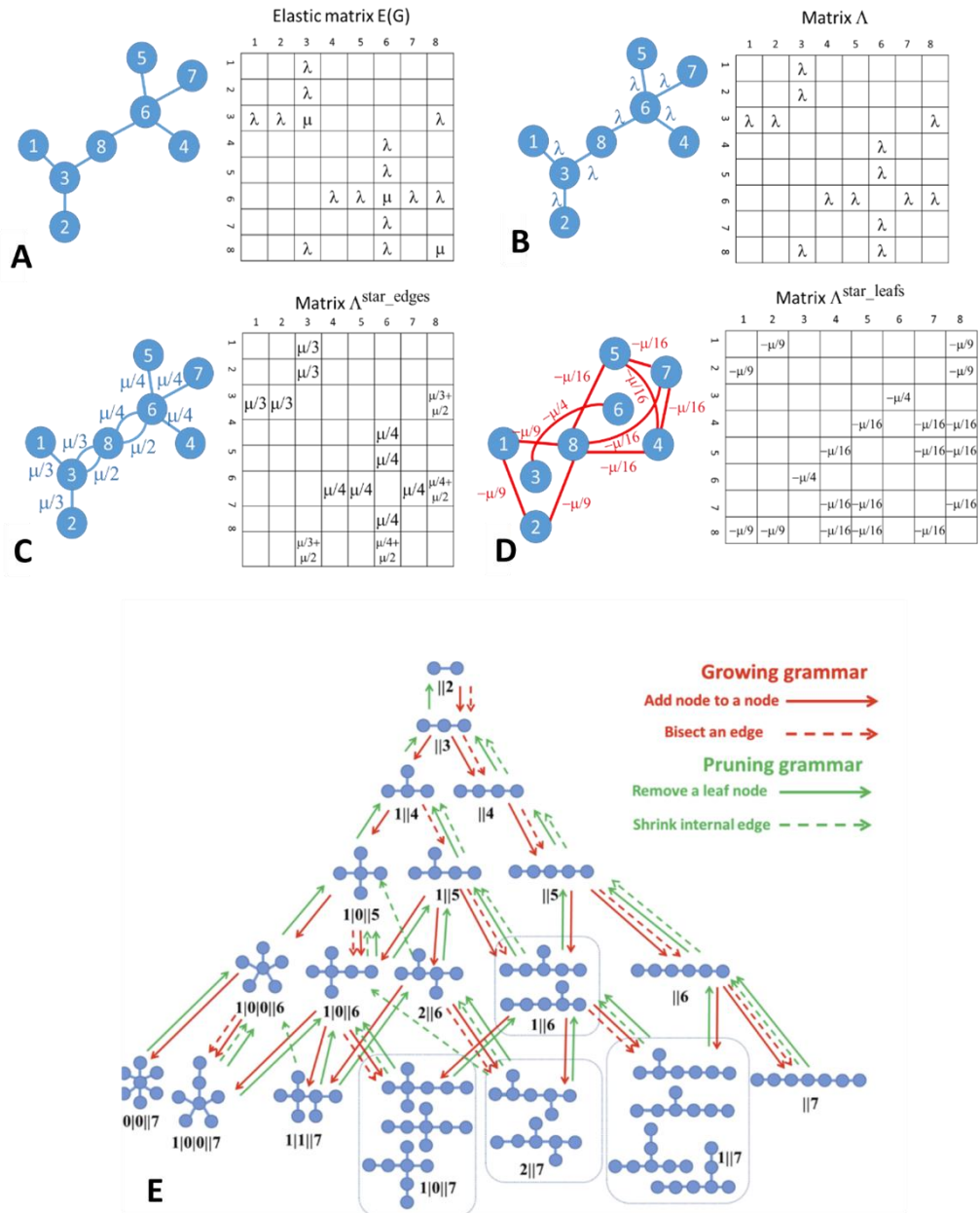

**Supplementary Figure 2. Elastic matrix-based definition of graph elastic energy and searching for the optimal graph topology in the structure space. (A-D)** The elastic matrix has dimension N x N, where N is the number of graph nodes (8 in this case). The stretching elasticity moduli $\lambda$ appears at the intersection of rows and columns, corresponding to each edge (weighted adjacency matrix, shown in panel B). The bending elasticity modulus $\mu$ appear at the diagonal elements of the matrix corresponding to the centers of graph $k$-stars. The bending elastic energy of the graph is described by two weighted adjacency matrices: with positive weights, where each edge receives a weight $\mu/k$ from each $k$-star to which it belongs (panel C), and with negative weights, corresponding to all possible pairwise connections between the leafs of each $k$-star, with weights $-\mu/k^2$ (panel D). **(E)** All the possible distinct tree-like topologies are shown for graphs with a number of nodes between 1 and 7. Each arrow illustrates the application of a graph rewriting rule (graph grammar operations). A rule can increase the number of nodes (growing, shown in red) or decrease the number of nodes (pruning, shown in green). ElPiGraph explores the space of structures starting from an initial node on this graph and following a trajectory determined by the local decrease of the elastic energy of the graph embedding into the data space. In the standard strategy, two growing operations are followed by one pruning operation, in order to avoid an irreversible trapping into a suboptimal graph structure.

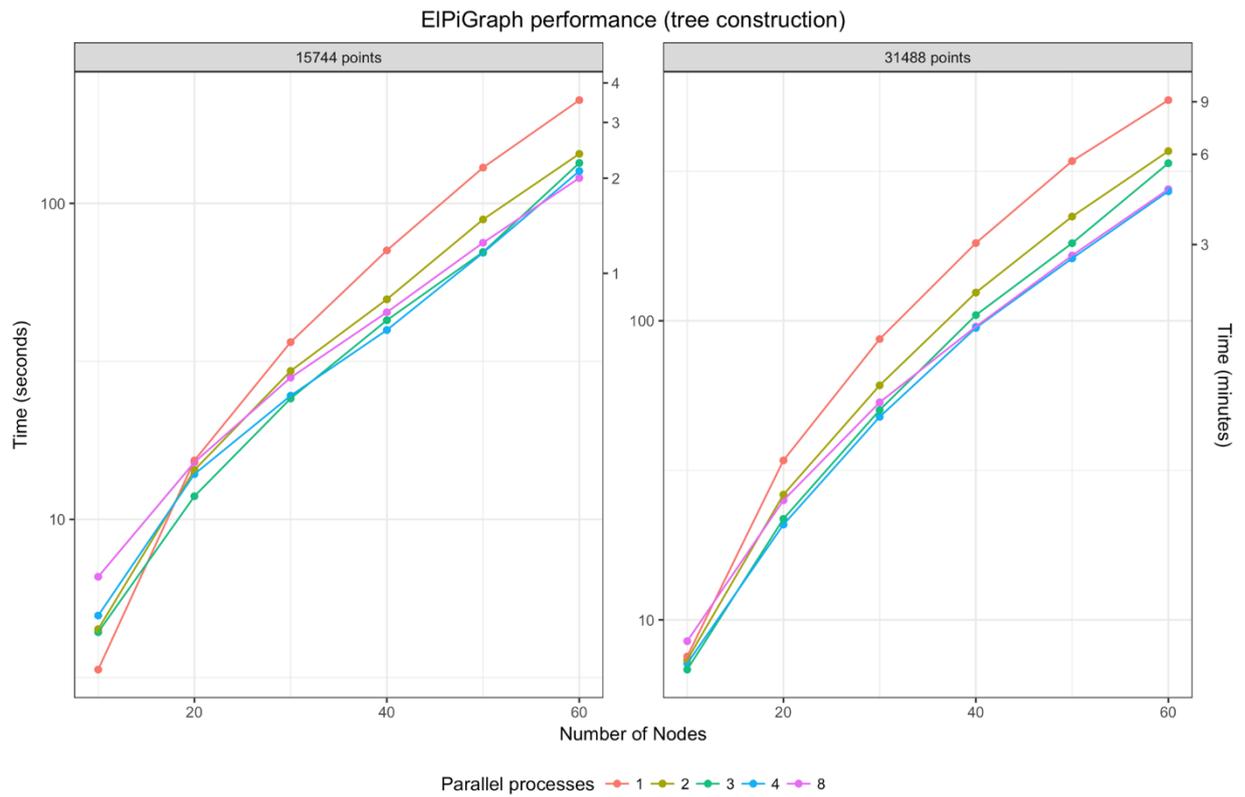

**Supplementary Figure 3.** Time employed by ElPiGraph.R for th reconstruction of a principal tree on a 3D datasets with ~15k points and ~31K points and different number of parallel processors under the same conditions of Figure 3.

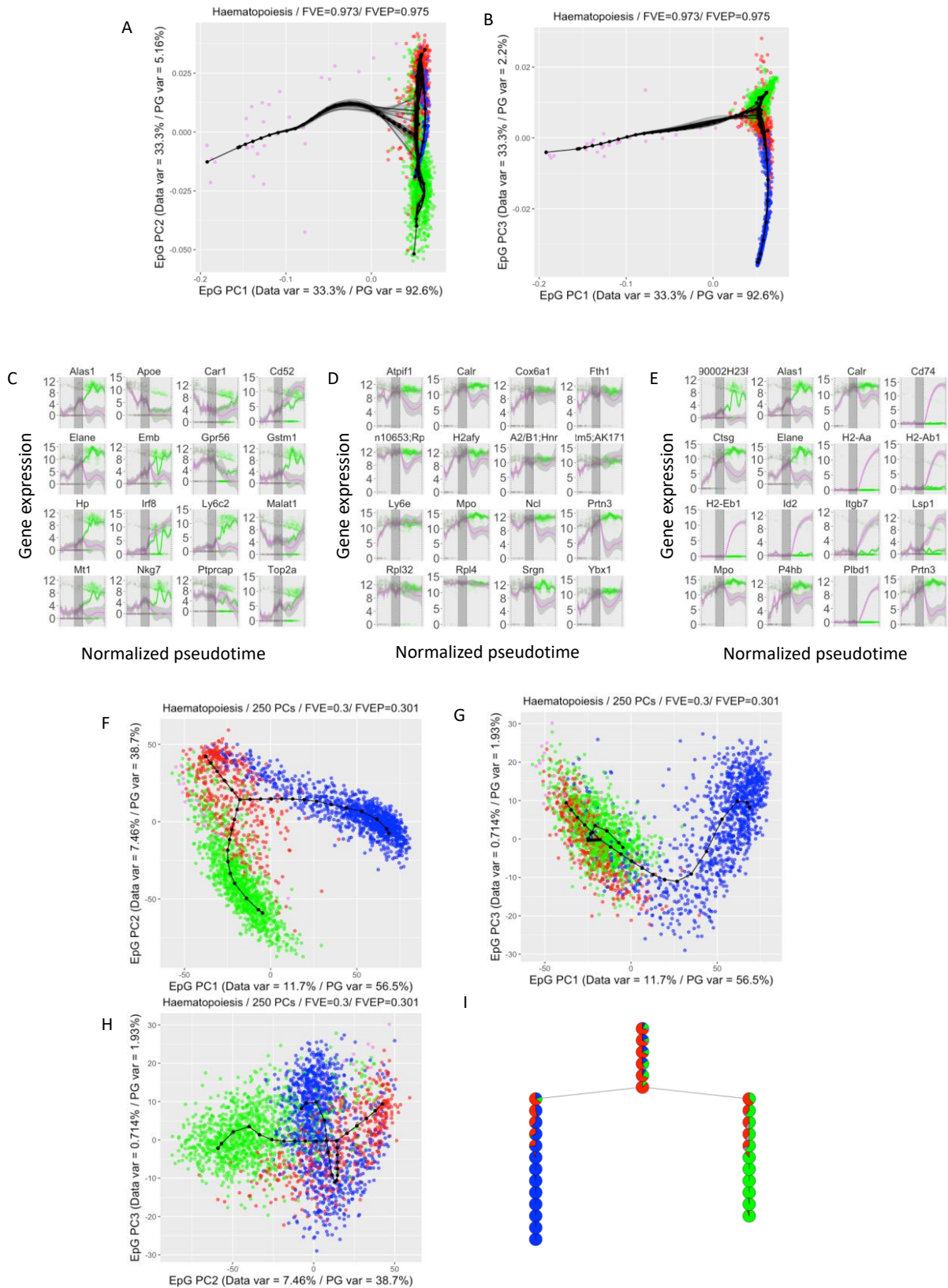

**Supplementary Figure 4. (A-B)** Projection of the hematopoietic data discussed in the main text and the principal trees constructed over them in to the 1st and 2nd and 1st and 3rd ElPiGraph principal components with the same graphical conventions of figure 4B. **(C-E)** Pseudotime-associated dynamics of genes selected using different criteria with the same convections of Figure 4D. **(F-H)** Projection of the hematopoietic data discussed in the main text without the application of MLLE and the principal trees constructed over them using the principal components derived from the data with the same graphical conventions of figure 4B. **(I)** Diagrammatic representation of the distribution of cells across the branches of the tree reconstructed by ElPiGraph on the non-MLLE transformed data. In all the panel, the same conventions of Figure 5 have been used.

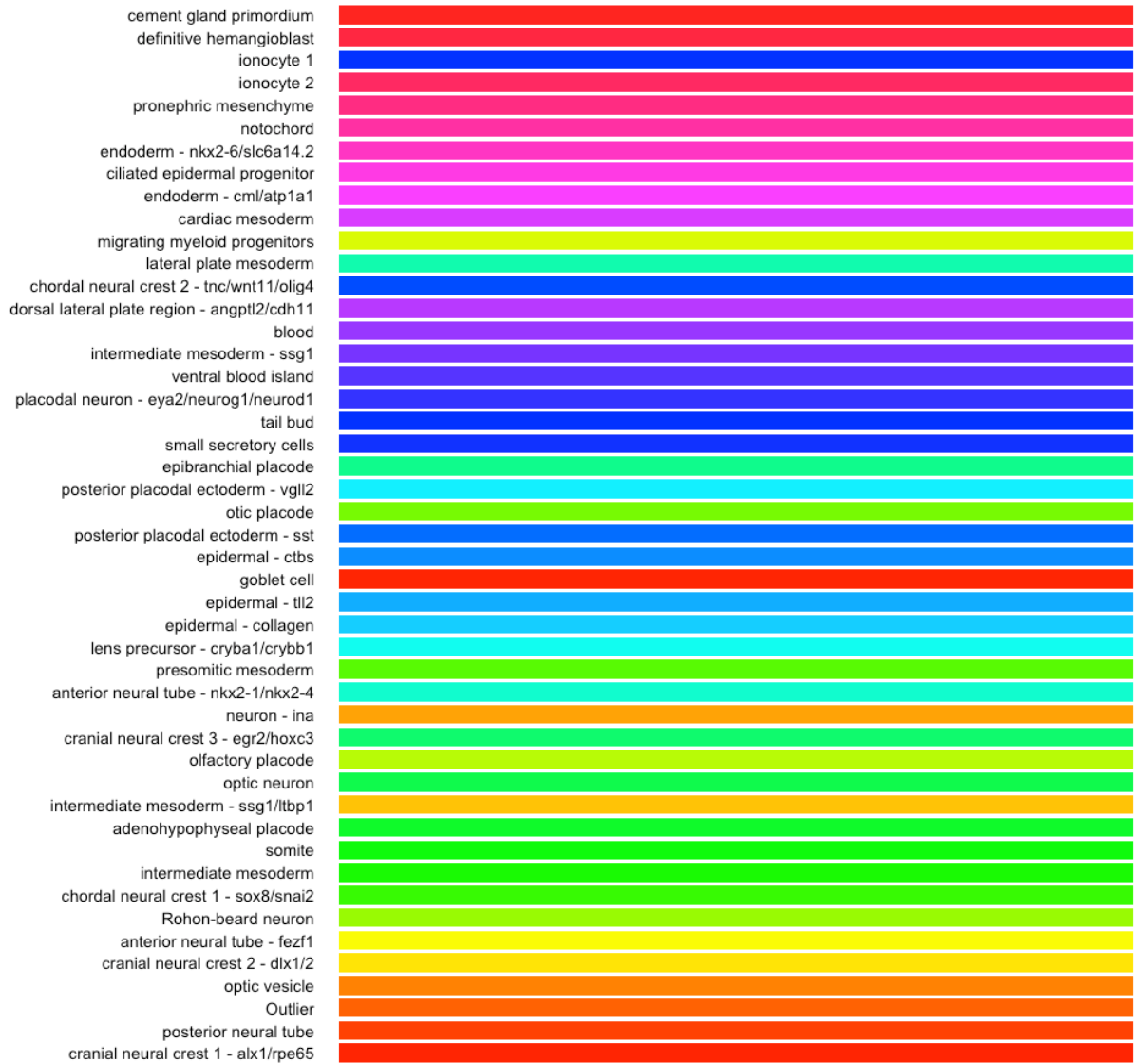

**Supplementary Figure 5. Color scheme for Xenopus.** The figure illustrates how different colors are used to indicates different cell type in the Xenopus embryo. The data have been obtained from the publication referenced in the main text.

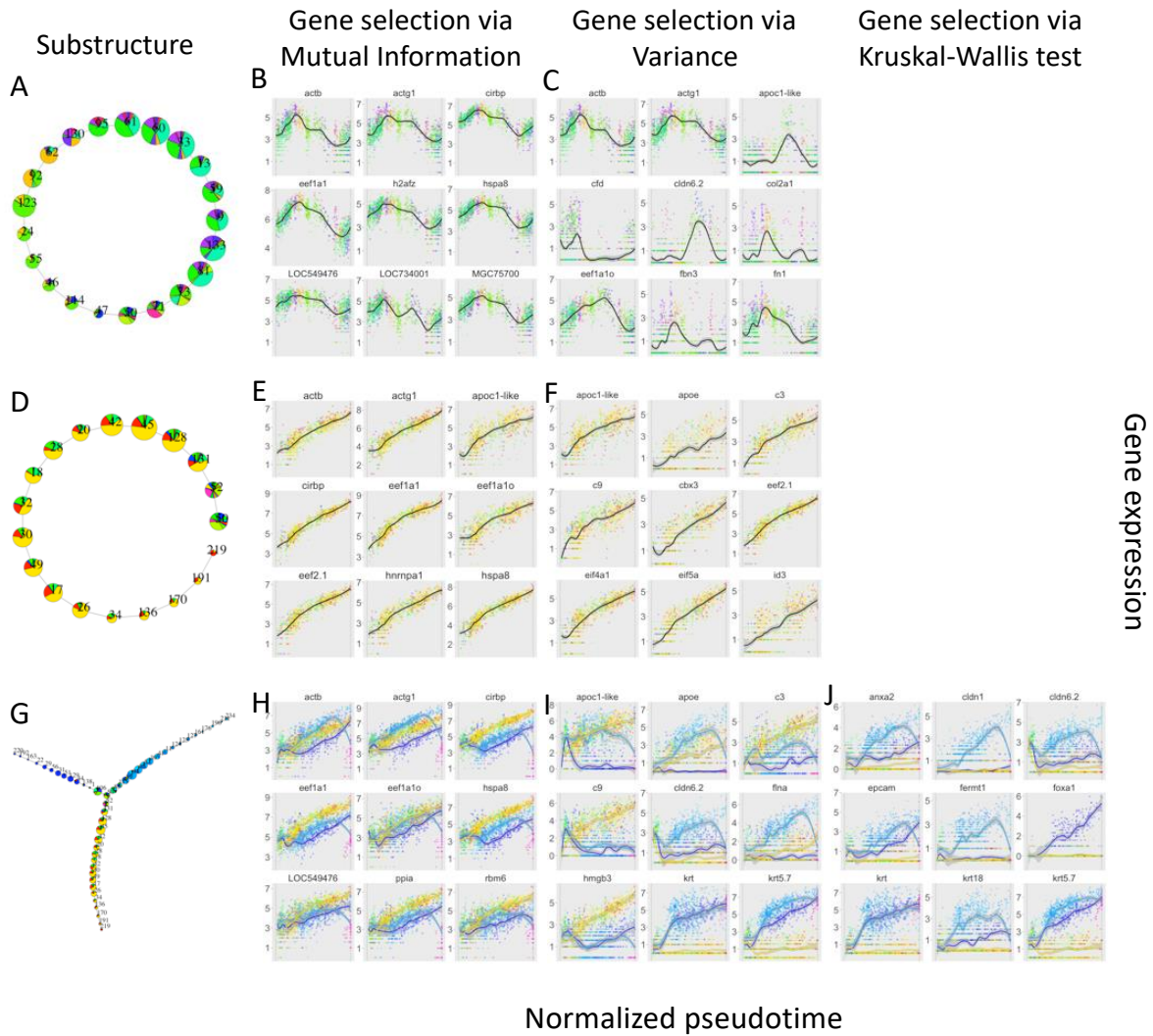

**Supplementary Figure 6. Additional reconstructed gene dynamics for the Xenopus embryo.** Three different substructures of the principal graph reported by Figure 5F (**A**, **D**, **G**) have been used to derive a pseudotime, and gene selection has been performed in ElPiGraph.R using mutual information (**B**, **E**, **H**), Variance (**C**, **F**, **I**), and the p value of a Kruskal-Wallis test when applicable (**J**). In all the relevant panels, points indicate gene expression in the cells and are colored to indicate the cell type, and a LOESS smoother with a 95% confidence interval has been used to fit the gene expression across the paths. In panels H-J, the color of the fitted smoother indicates the predominant cell population on the path.

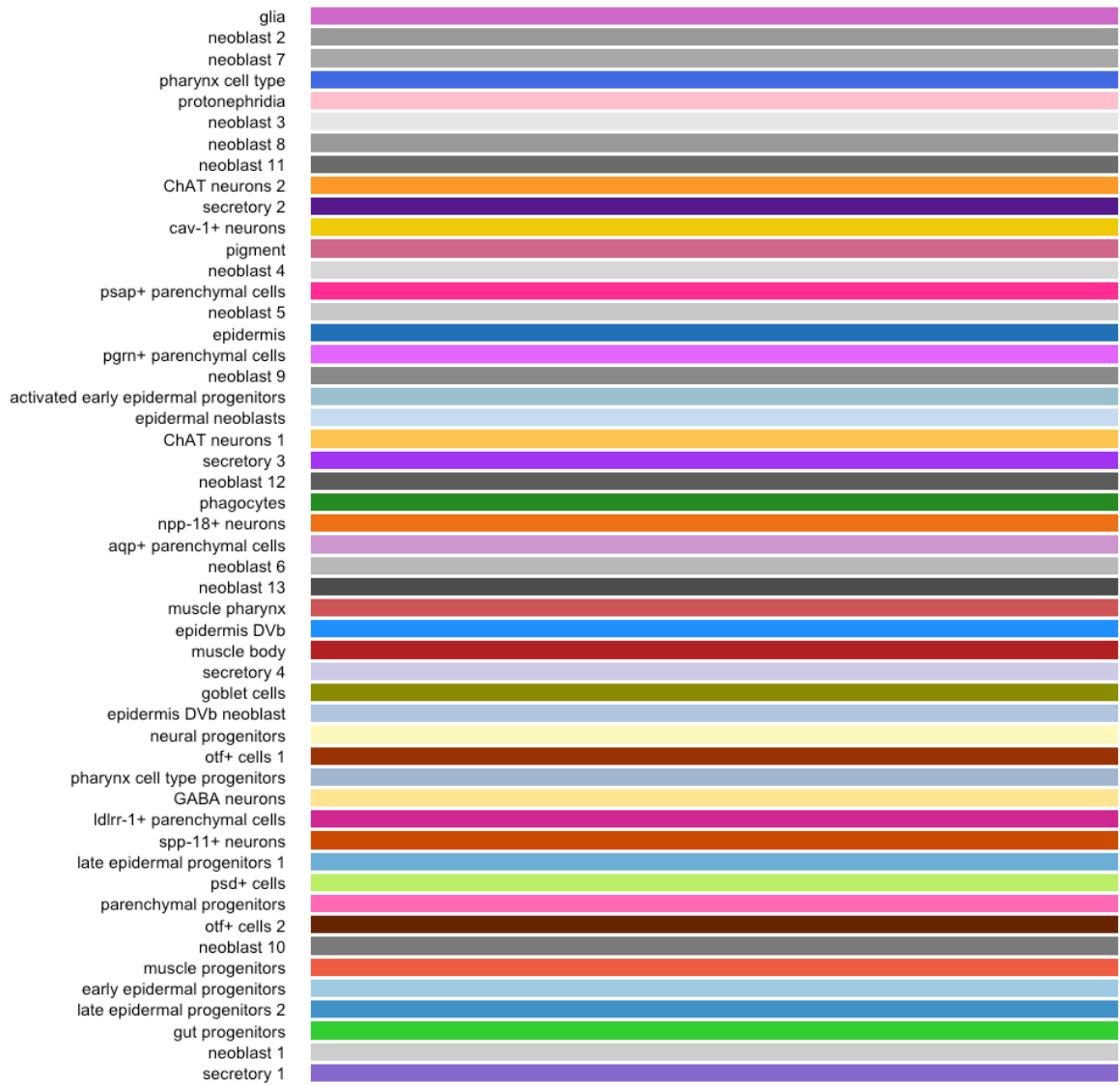

**Supplementary Figure 7. Color scheme for Planarian.** The figure illustrates how different colors are used to indicates different cell type in the Planarian data. The population information and color have been obtained from the publication referenced in the main text.

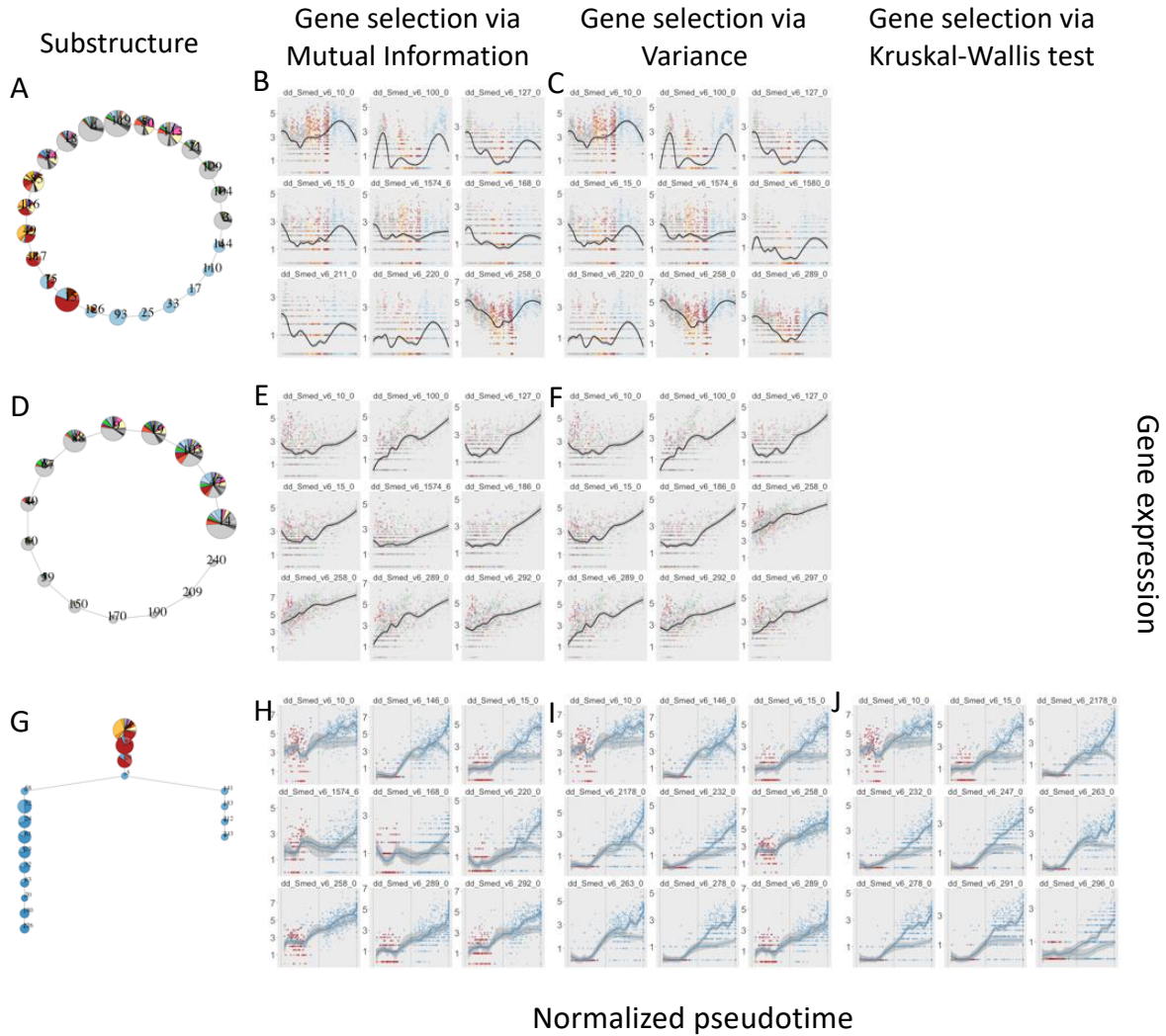

**Supplementary Figure 8. Additional reconstructed gene dynamics for Planarian.** Three different substructures of the principal graph reported by Figure 5I (**A**, **D**, **G**) have been used to derive a pseudotime, and gene selection has been performed in ElPiGraph.R using mutual information (**B**, **E**, **H**), Variance (**C**, **F**, **I**), and the p value of a Kruskal-Wallis test when applicable (**J**). In all the relevant panels, points indicate gene expression in the cells and are colored to indicate the cell type, and a LOESS smoother with a 95% confidence interval has been used to fit the gene expression across the paths. In panels H-J, the color of the fitted smoother indicates the predominant cell population on the path.